\documentclass{article}
\usepackage{ijcai24}

\usepackage{times}
\usepackage{soul}
\usepackage{url}
\usepackage[hidelinks]{hyperref}
\usepackage[utf8]{inputenc}
\usepackage[small]{caption}
\usepackage{graphicx}
\usepackage{amsmath}
\usepackage{amsthm}
\usepackage{booktabs}
\usepackage{algorithm}
\usepackage[switch]{lineno}

\usepackage[dvipsnames]{xcolor}
\usepackage{subcaption}
\usepackage{mathtools}
\usepackage{makecell}
\usepackage{enumitem}
\usepackage{amsfonts,amssymb}
\usepackage{bm}
\usepackage{multirow}
\usepackage{algorithmicx}
\usepackage{algpseudocode}
\usepackage{varwidth}
\usepackage{cleveref}
\crefformat{footnote}{#2\footnotemark[#1]#3}

\makeatletter
\newcommand{\thickhline}{%
    \noalign {\ifnum 0=`}\fi \hrule height 1pt
    \futurelet \reserved@a \@xhline
}
\newcolumntype{"}{@{\hskip\tabcolsep\vrule width 1pt\hskip\tabcolsep}}
\makeatother
\makeatletter
\newcommand{\labelname}[1]{% \labelname{<stuff>}
  \def\@currentlabelname{#1}}%
\makeatother
\title{A Novel GAN Approach to  Augment Limited Tabular Data for  Short-Term Substance Use Prediction}
\author{
Nguyen Thach$^1$
\and
Patrick Habecker$^1$\and
Bergen Johnston$^1$\and
Lillianna Cervantes$^1$\and
Anika Eisenbraun$^1$\and
Alex Mason$^1$\and
Kimberly Tyler$^1$\and
Bilal Khan$^2$\And
Hau Chan$^1$
\affiliations
$^1$University of Nebraska-Lincoln\\
$^2$Lehigh University
\emails
nate.thach@huskers.unl.edu,
\{phabecker2, bergenjohnston, lcervantes4, aeisenbraun3, amason19, ktyler2, hchan3\}@unl.edu,
bik221@lehigh.edu
}
\begin{document}

\maketitle

\begin{abstract}
    Substance use is a global issue that negatively impacts millions of persons who use drugs (PWUDs). In practice,
    % minimizing its potential harms by identifying vulnerable PWUDs for efficient allocation of appropriate resources is more effective yet often overlooked,
    identifying vulnerable PWUDs for efficient allocation of appropriate resources is challenging due to their complex use patterns (e.g., their tendency to change usage within months) and the high acquisition costs for collecting PWUD-focused substance use data. 
    Thus, there has been a paucity of machine learning models for accurately predicting short-term substance use behaviors of PWUDs. 
    In this paper, using longitudinal survey data of 258 PWUDs in the U.S. Great Plains collected by our team, we design a novel GAN that deals with high-dimensional low-sample-size tabular data and survey skip logic to augment existing data to improve classification models' prediction on (A) whether the PWUDs would increase usage and (B) at which ordinal frequency they would use a particular drug within the next 12 months.
    Our evaluation results show that, when trained on augmented data from our proposed GAN, the classification models improve their predictive performance (AUROC) by up to 13.4\% in Problem (A) and 15.8\% in Problem (B) for usage of marijuana, meth, amphetamines, and cocaine, which outperform state-of-the-art generative models.
    %\hau{try to see if we should mention drugs here using a few words}
    % \nateedit{how deep generative modeling can help improving predictive performance on similar domain problems that often deal with \emph{limited} and complex tabular (esp. \emph{survey}) data?}
    % The {\it IJCAI--24 Proceedings} will be printed from electronic
    % manuscripts submitted by the authors. The electronic manuscript will
    % also be included in the online version of the proceedings. This paper
    % provides the style instructions.
    % \hau{the title looks a bit too long}
    
    % \naterevise{Short-Term Substance Use Prediction on Limited Longitudinal Survey Data: A Novel GAN Approach}
    
    % \hau{in the submission, they are asking for additional information; please check and incorporate into the introduction}
\end{abstract}

\section{Introduction}

Substance use can create short- and long-term negative consequences for persons who use drugs (PWUDs) \cite{drug-report1,drug-report2,drug-report3}.
These consequences include mental illness, HIV/AIDS, hepatitis, drug overdose, and death.  
% \hau{I remember there is a sentence about over 90,000 deaths from drug overdose from some data source in 2020 or 2021??; add a sentence like this here}
% According to a 2021 National Survey on Drug Use and Health (NSDUH) from the Substance Abuse and Mental Health Services Administration (SAMHSA) \cite{SAMHSA2021report},
Within the U.S. alone, an estimated 161.8 million people aged 12 or older used a substance (out of which %54.7 million used tobacco, 133.1 million used alcohol, 31.6 million used marijuana, and 
40.0 million used an illicit drug) in the past month before being interviewed in 2021 \cite{SAMHSA2021report}.
Furthermore, according to \cite{NIDAdeathrates}, substance-involved overdose deaths, including those related to illicit drugs and prescription opioids, continue to increase over the years, with 106,699 deaths in 2021 compared to 91,799 (+16\%) in 2020 and 70,630 (+51\%) in 2019.
This alarming trend also applies globally, with the estimated number of PWUDs in the past 12 months reaching 296 million in 2021 from 240 million in 2011 (out of which 39.5 million and 27.3 million had drug use disorders, respectively) \cite{unodc2023}.

% \hau{add another sentence say/cite something about the trend continuing to increase in the future in the US} 

% \hau{add a sentence about they are not only an issue in the US but also a problem in the world; find/cite some statistics about drug use in the world and negative impact; costs; deaths etc... to explicitly show it is an indeed global issue}

% \hau{i have a paper about fighting forest fires; take a look the intro there}

In response to these large-scale negative impacts, various public and private organizations around the globe have prompted initiatives
% to examine the complex dynamics of substance use
for preventing and reducing substance use at both population and individual levels.
Prominent examples are the United Nation’s Sustainability Goal 3: ``Strengthen the prevention and treatment of substance abuse" \cite{United-Nations:sdg3} and U.S. Department of Health and Human Services (HHS)'s Healthy People 2030: ``Reduce misuse of drugs and alcohol" \cite{HHS:healthy-ppl30}.

% Combining with its complex dynamics \cite{diclemente2018addiction} \hau{what complex dynamics?} and global prevalence \hau{in terms of?} \cite{SAMHSA2021report}, the negative impact of substance use incurs astronomical economic and societal costs \hau{again you have to at least mention them to show it is very huge; is it for US only? which countries? for what drugs?} \cite{RCA2019economic} and prompts United Nation’s Sustainability Goal 3 \cite{United-Nations:sdg3}: ``Strengthen the prevention and treatment of substance abuse".

% \hau{I think you want to start the above paragraph to say something like because of these negative impacts around the globals, various private and public organizations have designed initiatives to examine complex dynamics ??? in order to prevent and reduce substance uses at various levels. These initiatives include ???, United Nation’s Sustainability Goal 3 (i.e., `Strengthen the prevention and treatment of substance abuse") mention a few things here ... in the submission they ask for something, it can go here it think} 

% \hau{below make connections to the above paragraph; check whether the following is true}

Generally, approaches toward these initiatives focus on designing and deploying intervention and outreach programs/resources (e.g., rehabs and consulting services) for PWUDs, with the main goal of reducing and eliminating their usage of certain substances \cite{ray2020combined,colledge2023global}.
While these programs and resources have shown to be effective to some extent \cite{ouimette1997twelve,tanner2016adolescent}, they often require volunteer participation from PWUDs, who face the difficulties and reluctances of (self-)evaluating and (self-)determining whether they want or need help \cite{russell2021qualitative,wilson2023barriers}. 
%By the time 
Even when PWUDs agree to use these programs/resources, they may have already experienced prior harms such as overdose and mental illness \cite{SAMHSAAdmissions,andersson2023inpatients}. 
Therefore, it is important to prevent harm from occurring in the first place by carefully identifying PWUDs at the highest risk (i.e., those who are prone to drastically increase usage of some drug) and allocating them appropriate resources to reduce or eliminate potential harms. 

Unfortunately, forecasting individual substance use behavior is challenging due to its complex patterns (e.g., drug use frequency and co-use of multiple drugs) and tendency to change over time at short timescales (i.e., within months) \cite{karamouzian2022latent,linden2022stress,lorvick2023protocol},
% \hau{we need to add a sentence and cite something about PWUDs usage change over time and it can be a challenge to predict the changes? if you say over time is short-term (within months), then i think we can make some argument that existing data/models don't apply in the limitation paragraph below?}
as well as the lack of appropriate data and models for predicting short-term future substance use (see \textbf{Existing Efforts and Limitations} and additional contents in the full paper attached in the supplementary material). 

%\hau{i am not sure where this paragraph would go}
Thus, our goal is to design accurate predictive models for modeling short-term drug usage (i.e., within months) from PWUDs to aid healthcare agencies, local communities, policymakers, and other stakeholders in the efficient allocation of resources to PWUDs who need the most help.
An effective predictive model has the potential to improve the well-being of millions of PWUDs and hampers the ongoing rapid growth of drug use prevalence and drug overdose death rates. %\cite{NIDAdeathrates}. 

\paragraph{Our Approach and Associated Challenges.}
To address the above shortcomings of substance use data and models,
we formed a collaboration between computer scientists and domain experts in substance use research, including social scientists, intervention specialists, and survey interviewers from University of Nebraska-Lincoln and Rural Drug Addiction Research Center.
With IRB approval, our team recruited a local sample of 258 PWUDs in the Great Plains of the U.S. from which longitudinal survey data were collected and stored in tabular format (detailed below in Section \ref{subsec:bg}).
Despite its high value and relevance to our stated goal, the sample size is small due to various challenges during data collection (mainly disruptions due to COVID-19 and the lack of visible harm reduction movements and punitive laws in the region),
% which in turn took us considerable time to find and recruit PWUDs for the sample and to gain their trust).
As a result, our preliminary predictive models trained on currently available data achieved subpar performance with respect to the baseline (demonstrated in our full paper). %(discussed in Section \ref{subsec:chal}).
These models seem to overfit due to the limited training examples ($<$200) with respect to the number of (preprocessed) features ($>$600). %\hau{due to ???}. 
% \hau{IRB-approved or something should go somewhere in this paragraph??}

% \hau{All submissions should clearly state WHAT real-world problem is being tackled, WHO is participating in the project / paper (such as AI grassroot movement organizations, entrepreneurs, policy makers, community leaders, non-profits and universities), HOW is the topic being handled and the results are measured (metrics). When we say our team, maybe we need to say may be social scientists, substance use experts, intervention persons (look up Alex Mason at UNL)?, we need to think about this; also check with Kim}

One effective way to tackle overfitting on small datasets %\hau{so all we doing is to limit overfitting? or do you mean due to limited sample size?}
is via data augmentation \cite{shorten2019survey} i.e., creating synthetic samples based on real data to increase its sample size.
\emph{Deep generative models} such as the popular Generative Adversarial Network (GAN) \cite{goodfellow2014generative} provide powerful tools for this purpose due to their flexibility in representing complex and high-dimensional data distributions. 
%Therefore, they have been successfully applied to various domains in computer vision and natural language processing \cite{nikolenko2021synthetic}.
In recent years, these models have been extended to generate tabular data, especially healthcare records, using specialized architectures (e.g., GOGGLE \cite{liu2022goggle}) and/or novel training algorithms (CTGAN \cite{xu2019modeling}), which in turn achieved promising performance across different evaluation criteria.
% This particular data type is of great interest due to its heterogeneity (containing different feature types, e.g., categorical and continuous as well as missing values) and often high dimensionality with imbalanced categorical features. 
% Furthermore, the correlation between features is either partially observed or unknown. Hence, conventional inductive biases for learning spatial or semantic relationships cannot be applied similarly as with image or natural language data \cite{borisov2022deep}.
% State-of-the-art deep generative models attempted to tackle these challenges using specialized architectures (e.g., GOGGLE \cite{liu2022goggle}) and/or novel training algorithms (CTGAN \cite{xu2019modeling}), which in turn achieved promising performance across different evaluation criteria (e.g., likelihood fitness and how indistinguishable synthetic samples are from real ones judging by some external classifier). 
%judging by some classifier trained on real data)}. %\hau{(e.g., list one or two)}.

However, these models were benchmarked mostly on moderate to large datasets that typically contain at least thousands of training examples.
In rare cases when being applied on small datasets with less than 1000 samples, e.g., \href{https://www.openml.org/search?type=data&sort=runs&id=37&status=active}{Diabetes} (768) and Breast (569) \cite{street1993nuclear} as reported respectively in the work for GOGGLE and TabDDPM \cite{kotelnikov2023tabddpm}, the proposed models and its selected comparatives often either fail to or narrowly outperform simple baselines such as Bayesian networks \cite{pearl2011bayesian} and SMOTE \cite{chawla2002smote} in terms of their considered evaluation metrics. %\hau{outperform in what sense? why ain't other baselines like LG or Trees being considered?}
The benchmark datasets for state-of-the-art models also contain no more than 200 features, which is substantially smaller than the size of our feature set ($>$600 after preprocessing).
% In their discussion, the authors of GOGGLE explicitly stated that their model mainly considers settings with less than 100 features, which they deemed reasonable for real-world tabular data.
% Since we are working with a substantially larger feature set, these models naturally do not perform well as demonstrated shortly in our experiments.
Additionally, the survey used for collecting our tabular data contains \emph{skip logic}, a commonly employed functionality in survey design for social science applications \cite{fowler1995improving,dillman2014internet}, which has not been addressed by the aforementioned state-of-the-art models.

% \hau{we need to add a sentence about skip logic here and contribution; it seems that above argument is for small data; we also need to add a sentence about they don't work well for our settings additionally because our data contain skip logic or something like that? }

\paragraph{Our Contributions.}
In this paper, we design a novel specialized generative model to augment our tabularized survey data with skip logic in order to improve the predictive performance of our classification models, which ultimately predict the following two short-term substance use behaviors: for a given PWUD and a certain drug, (A) whether they would increase its usage and (B) at which frequency (on a pre-defined ordinal scale) they would use it within the next 12 months. 
We summarize our contributions as follows: 

\begin{enumerate}[label=(\Roman*)]
    \item We believe ours is the first work that addresses \emph{skip logic} from surveys in tabular data generation. We demonstrate its practical value by showing both conceptually and empirically how enforcing constraints from skip logic (or \emph{skip constraints} for brevity) positively affects the training of our generative model.
    We are also one of the first to investigate the real-world feasibility of deep generative models in settings where the number of features exceed the sample size i.e., high-dimension low-sample-size (HDLSS). (See \textbf{Related Work} in our full paper.)
    \item We design a novel GAN that deals with small tabular data containing $<$258 samples and 210 features (209 plus one target variable). Specifically, we leverage CTGAN \cite{xu2019modeling}, a well-known tabular GAN, by incorporating an auxiliary classifier \cite{park2018data,zhao2022ctab} within its architecture to generate high-quality samples conditioned on the corresponding target variable. Since the transformed feature space has high dimensionality (over 600), we also embed a global feature selection mechanism while training the auxiliary classifier's by employing the novel approach from \cite{margeloiu2023weight} that was shown to perform well on even smaller and higher dimensional data than ours. Finally, to enforce the skip constraints stated in (i), we take advantage of CTGAN's built-in conditional vector. 
    % the reduced total number of learnable parameters from WPFS balances out the additional time to train WPN and SPN, hence there is no noticeable difference in runtime between the models trained with and without its addition.
    \item We implement and train the proposed GAN and use it to augment our (training) data. 
    The augmented data is then used to train binary and multiclass classification models for predictive problems (A) and (B), respectively, for each of the following drugs: marijuana, methamphetamine, amphetamines, and cocaine. Our experimental results show that the average Area under the Receiver Operating Characteristic curve (AUROC) evaluated on multiple distinct sets of test data is improved by up to 13.4\% in Problem (A) and 15.8\% in Problem (B) when the data is augmented, which is significantly higher than what yielded using state-of-the-art generative models. %\hau{we didn't say what drugs here?}
\end{enumerate}

% The remainder of the paper is outlined as follows. We provide the preliminaries, problem formulation, and motivations for our proposed GAN in Section \ref{sec:prob-desc}.
% We elaborate on the proposed GAN's design in Section \ref{sec:gan} and rigorously benchmark its performance using our proposed evaluation framework in Section \ref{sec:exp}.
% % Additional related work can be found in Section \ref{sec:related}.
% We conclude our work in Section \ref{sec:conclu}.
% The full paper can be found in the supplementary material.
% \hau{might need to say something about appendix or supplementary if we are cutting something out later}

% \hau{include outline of the sections of the paper}

% United Nation’s Sustainability Goal 3 \cite{United-Nations:sdg3}: ``Strengthen the prevention and treatment of substance abuse"
% \begin{itemize}
%     \item substance use is global with highly dynamic and complex patterns
%     \item in response, we conduct a longitudinal study using a small local sample from rural areas of the U.S. Great Plains.
%     \item predictions are difficult given the nature of the data (tabular and often HDLSS) $\longrightarrow$ we run several state-of-the-art models on our data
%     \item 
% \end{itemize}

% Our contributions are as follows:
% \begin{enumerate}
%     \item 
% \end{enumerate}

% The remaining of the paper is outlined as follows.

\section{Problem Description}\label{sec:prob-desc}

% \hau{there should be some texts here; maybe talk about what you are going to do and what at each subsection}

% We first describe the characteristics of our tabular data and defining relevant concepts (Section \ref{subsec:bg}), then formulate our problems of interest (Section \ref{subsec:prob-formu}), and lastly conduct preliminary experiments to motivate the design of our proposed GAN (Section \ref{subsec:chal}).

\subsection{Background}\label{subsec:bg}

% \hau{I think we should add a footnote about the availability of the data Patrick sent and anonymize it}

\paragraph{Our Tabular Data.}

The recruitment of PWUDs started in 2019 under the respondent-driven sampling scheme \cite{heckathornRDS} in the Great Plains of the U.S. and has continued to the present time. Enrolled PWUDs were followed up within 4--12 months after their initial visit and took the same survey as before.
In the survey, each PWUD answered questions on a computer regarding their individual attributes, including the drug use behavior of 18 different drugs. % in a largely self-administered manner.
Use frequencies of considered drugs (e.g., marijuana, cocaine, amphetamines, and methamphetamine) were inquired on an ordinal scale (1--8) of \{never, less than once a month, once a month, once a week, 2--6 times a week, once a day, 2--3 times a day, 4 or more times a day\}.
%of 8 injection drugs (e.g., injection heroin, injection opioids, and injection methamphetamine) %injection cocaine, heroin-cocaine %speedball, heroin-meth speedball, crack cocaine, and buprenorphine) 
%and 10 non-injection drugs (e.g., marijuana, cocaine,
% Ecstasy, PCP,
%amphetamines, and methamphetamine) 
% barbiturates, benzodiazepines, opioids, and heroin) 
%was inquired on an ordinal scale (1-8) of \{never, less than once a month, once a month, once a week, 2-6 times a week, once a day, 2-3 times a day, 4 or more times a day\}. %\hau{are you going to mention the non-injection drugs/scales? why did you take them out?} % for non-injection drugs, and on a similar ordinal scale but with one extra category of “never injected” added prior to “never” for injection drugs (to be discussed shortly).
Collectively, the responses to these questions form the (raw) features stored in a 2-D table.%\footnote{Due to the confidentiality agreement and IRB approval for this study, we do not have access to sensitive information such as full name and gender identity. Data will be made available to applicants (with an IRB protocol and ethical research plan) upon request.} 
%The applicants are required to have an IRB protocol.}. 
% prevent deductive identification and ensure that the center’s ethical obligation to study participants are met
%\hau{rows and columns? rows are what columns are what?}.
% \hau{this sentence might not be important; footnote it or something} 
There are 258 samples (each represented as a row) and 151 features (each represented as a column) in total.
(See supplementary material for the survey questions.)
%Nearly all features are categorical. 
%, which were already label encoded beforehand.
%Continuous features, e.g., age, are recorded in years and hence are nonnegative.
After being preprocessed (detailed in Section \ref{subsec:method}), the table contains 209 features (plus the target variable for teaching the desired classification models): 2 continuous and 207 categorical. %(out of which 138 are binary, 48 are ordered, i.e., taking values in a finite ordered set, and 21 are unordered).

% \hau{what does ordered or unordered mean here? maybe they are standard? people know them?}
% \hau{you need a sentence to say what are you doing next; you cannot just state something without saying stuff}

\paragraph{Definition 1 (Skip Logic).}\labelname{Definition 1}\label{def:skl}
In survey design, skip logic
% (also called conditional logic, branching, routing, filters, or simply skips)
is a set of automated navigational rules that allows respondents to skip to relevant questions depending on their prior answers \cite{couper2008designing}. %, which is especially helpful for long surveys.
% Figure \ref{fig:skl} gives an example of skip logic in our survey.
Each of the rules is defined as a
% constraint called 
\emph{skip constraint}, which restricts the possible values of a subset of features $A$ in accordance with the value of some feature $a$. We say the skip constraint on the \emph{chain} $A$ is \emph{imposed} by $a$ and denote this as $a\rightarrow A$. For the example in Figure \ref{fig:skl}, the skip constraint on $A=$ \{TB4\} is imposed by $a=$ TB3 i.e., TB3 $\rightarrow$ \{TB4\}, in which TB4 is \emph{omissible} when TB3=``No".

% We discuss in detail the challenges associated with the presence of skip logic within the context of predictive modeling shortly.
% More specifically, the presence of skip logic disrupts data statistics

\vspace{-2mm}
\begin{figure}[h]
    \centering
    \begin{minipage}{.45\textwidth}
    \scriptsize
    \fbox{\textbf{TB3}}\; \textbf{In the past 6 months, have you smoked cigarettes?}
    \begin{enumerate}[noitemsep,leftmargin=4em,topsep=0pt]
        \item[] No (0) --- SKIP TO TB5
        \item[] Yes (1)
    \end{enumerate}
    \fbox{\textbf{TB4}}\; \textbf{How many cigarettes do you usually smoke in a day?}
    \begin{enumerate}[noitemsep,leftmargin=4em,topsep=0pt]
        \item[] Not at all (1)
        \item[] Less than 1 cigarette a day (2)
        \item[] 1-5 cigarettes a day (3)
        \item[] Half a pack a day (4)
        \item[] A pack or more a day (5)
    \end{enumerate}
    \end{minipage}
    \caption{Skip Logic: If respondents answer ``No" in TB3, they will automatically be directed to TB5 without being asked on TB4.}
    \label{fig:skl}
\end{figure}
\vspace{-2mm}

% \hau{insert something here what Definition 2's implications? why do you need to present this?}

% We now define the task of tabular data generation which would be followed when evaluating generative models.
\paragraph{Definition 2 (Tabular Data Generation).}\labelname{Definition 2}\label{def:tab-gen}
Following the notations from \cite{xu2019modeling}, given a table $\mathbf{T}_{real}$ that is partitioned row-wise into training set $\mathbf{T}_{train}$ and test set $\mathbf{T}_{test}$, the task involves training a data generator $G$ on $\mathbf{T}_{train}$ and then independently sampling rows using the learned $G$ to generate a synthetic table $\mathbf{T}_{syn}$ such that $|\mathbf{T}_{syn}|=|\mathbf{T}_{train}|$ with similar probability mass function for some target variable, $y$ (defined in Section \ref{subsec:prob-formu}) (in order for us to fairly evaluate the efficacy of $G$).

% \hau{should this paragraph be part of the definition?}

% The number of synthetic samples to generate must match the sample size of real training data and the distribution of classes between the two corresponding tables must match in order for us to fairly evaluate the efficacy of $G$ e.g., how useful $\mathbf{T}_{syn}$ can be for improving the predictive performance of classification models trained on $\mathbf{T}_{train}$ when predicting for $y$ in $\mathbf{T}_{test}$ (formally defined in Section \ref{subsec:method}).
Note that a capable $G$ can satisfy the latter requirement without affecting the overall quality of the generated synthetic samples---that is, the features in each generated row should be consistent with the label in that row.
% \hau{why do they need to have the same size? also what target variable y? maybe you mentioned above? but i am confused, you only care about y? so like if 0.4 is T and 0.6 is F, you can just generate something random without caring about the data quality of the other features?}
% and $P_{\mathbf{T}_{train}}(y=i)=P_{\mathbf{T}_{syn}}(y=i)$ for every class $i$ of the target variable $y$.
% Note that for our setting, $\mathbf{T}_{real}$ and its derivatives include $y$ and hence 
For our setting, $\mathbf{T}_{real}$ and its derivatives consist of $N_c=2$ continuous features and $N_d=207+1$ categorical features. 
% \hau{What is Nd d for?}
%For clarity, we use $\mathbf{X}$ instead of $\mathbf{T}$ to denote similar tables but without $y$. 

% \hau{in the experiments, did you try different samples of Tsyn since each Tsyn can be different from G? what if you have a different size of Tsyn?}

% \hau{so when you evaluate the performance using test data, you don't use the fake generated data right?} 

% \hau{insert something here what Definition 3's implications? why do you need to present this?}

% In the following, we introduce the concepts of GAN and its relevant extensions that constitute the foundation of our proposed GAN.

\paragraph{Definition 3 (GAN and Its Extensions).}

GANs are \emph{deep} generative models that have recently found success in modeling tabular data \cite{borisov2022deep} in addition to images and text.
A GAN typically consists of two separate networks: a generator $\mathcal{G}$ that maps a noise distribution (typically Gaussian) to the data distribution and a discriminator $\mathcal{D}$ that estimates the probability an input sample came from the data distribution. The learning process is defined as an adversarial game between $\mathcal{G}$ and $\mathcal{D}$ in which $\mathcal{G}$ attempts to consistently fool $\mathcal{D}$ \cite{goodfellow2014generative}. 

To stabilize the training of GANs, \cite{arjovsky2017wasserstein} introduce WGAN that provides a meaningful loss, the Wasserstein distance, for quantifying the difference between the generated and real data distributions.
% Its extension, WGAN-GP \cite{gulrajani2017improved}, enforces the Lipschitz constraint on $\mathcal{D}$ (for robustness guarantees) %\cite{pauli2021training}
% by incorporating a penalty on the gradient norm i.e., \emph{gradient penalty} (GP) in the value function, instead of using weight clipping as proposed in the original work of WGAN. Such an alternative further improves training stability and requires less hyperparameter tuning.
Using the value function from WGAN-GP \cite{gulrajani2017improved} (an improvement of WGAN), \cite{xu2019modeling} design CTGAN that aims to tackle class imbalance in categorical features of tabular data by modifying $\mathcal{G}$ to additionally take a vector as input.
% \cite{xu2019modeling} design a GAN for tabular data using the value function from WGAN-GP \cite{gulrajani2017improved} (an improvement of WGAN). The proposed model, CTGAN, tackles class imbalance in categorical features by modifying $\mathcal{G}$ to additionally take a vector as input. %, which is coined \emph{conditional generator}.
This so-called \emph{conditional vector} represents a certain class/category of some categorical feature and is used to condition both the generated samples and the real training samples. %\hau{is the conditional vector more compact? Do they only select some features?} 
The model can thus efficiently learn proper conditional distributions for each feature. We further describe the main components of CTGAN in Section \ref{subsec:cond-gen}.

% \hau{i wonder why you present definition 3; you should say why you are presenting these definitions; why are they important for the readers to know; how are they related to what you are proposing to do}

\subsection{Problem Formulation}\label{subsec:prob-formu}
In this work, we focus on \emph{data augmentation} by designing a novel GAN to generate high-quality synthetic samples that resemble our tabular data with small sample size and skip logic incorporated. %\hau{and add something skip logic? by creating high-quality synthetic samples that are similar to the real data??}. 
The augmented data is then used to train classifiers for predicting PWUDs' usage of a given drug within the next 12 months, including (A) whether they would increase its usage and (B) at which ordinal frequency they would use it. The former is a binary classification problem wherein PWUDs exhibiting an increase and non-increase (i.e., decrease or unchanged) in usage belong to the positive and negative class, respectively; the latter is a multiclass classification problem in which PWUDs are labeled according to their ordinal usage of the corresponding drug within 12 months after.
We mainly concern with predicting usage of the most prevalent drugs (i.e., used by at least half of the PWUDs in our tabular data), which include marijuana, methamphetamine (or meth for brevity), amphetamines, and cocaine.
% two predictive problems

There are two unique properties of our tabularized survey data
% However, our tabularized survey data have two unique properties
that impede existing generative and predictive models from achieving desirable performance: being high-dimension low-sample-size (HDLSS) and the presence of skip logic %from conventional predictive models. 
(whose impact is demonstrated in \textbf{Challenges and Observations} of the full paper). 

\section{Our Proposed GAN}\label{sec:gan}

% We start this section by providing an overview of our proposed GAN in Section \ref{subsec:overview}, followed by an in-depth description for each of its essential components: i) the conditional generator (Section \ref{subsec:cond-gen}), ii) the auxiliary classifier (Section \ref{subsec:ac}), iii) the built-in mechanism for identifying important features during GAN training (Section \ref{subsec:wpfs}), and iv) our technique for enforcing skip logic (Section \ref{subsec:skl-enforce}). We put together the complete model in Section \ref{subsec:complete-gan}.
% \hau{add overview sentences here}

% \begin{figure}
%     \centering
%     % \includegraphics{}
%     \caption{Caption}
%     \label{fig:gan-archi}
% \end{figure}

\subsection{Overview}\label{subsec:overview}

% Algorithm \ref{alg:1} summarizes our proposed GAN.
We focus on GAN architectures due to their prevalence in tabular data generation literature \cite{borisov2022deep} and their convenient properties (e.g., flexibility for conditional generation) that help us approach the discussed challenges systematically. %in Section \ref{subsec:chal}.
Overall, we extend the popular CTGAN (reviewed in Section \ref{subsec:cond-gen}) as follows:
% The key components are:

% The model's goal is to learn from HDLSS data and generate useful synthetic samples for downstream classification task.
\begin{enumerate}[label=\roman*)]
    \item Due to the large number of (categorical) features, it is difficult for the conditional generator $\mathcal{G}$ to learn to generate samples that are conditioned on a particular feature. 
    %(since during each update of the model's parameters, only [$batch\_size$] features are randomly selected upon which synthetic samples are conditioned).
    Therefore, during the training of $\mathcal{G}$, we raise the production of synthetic samples
    % by a factor of $q > 1$ with respect to the batch size ($q=1$ in CTGAN) 
    in order to ensure adequate training for conditional generation on a wide range of features.
    Furthermore, we prioritize the generation of synthetic samples that are conditioned on the empirical distributions of the target variable. %---that is, those being generated from the learned conditional distribution of rows given a specific class label.
    % \hau{why does this argument tie to stochastic gradient descent? could other methods work besides SGD? I think your argument shouldn't tie to this method only; what do you mean by conditioned on the target variable the last sentence here and next?}
    
    % Due to a large number of conditions (i.e., categorical columns) to select \hau{select for what? also what do you mean by conditions?}, it is difficult for $\mathcal{G}$ to learn to generate samples with the input condition preserved \hau{don't understand what you mean by input condition here}.
    % Therefore, during the training of $\mathcal{G}$, we increase the sample size of the conditions by a factor of $q$ with respect to the batch size ($q=1$ in CTGAN).
    % Furthermore, we dedicate a portion of the sample to conditions for the target variable only. \hau{I am not clear about the last two sentences; what impact of the q? also not clear about the portion of the sample to conditions ...}
    \item To enhance the quality of synthetic samples \emph{conditioned on} the target variable $y$, we incorporate an \emph{auxiliary classifier} $\mathcal{C}$ \cite{park2018data} into the architecture of CTGAN for learning the correlation between $y$ and other features. That is, $\mathcal{C}$ is trained to predict the label of an input (synthetic or real) sample with $y$ removed. %\hau{why do you need to do this? what if you don't?} 
    \item Within $\mathcal{C}$, which is originally a multilayer perceptron (MLP), we add two auxiliary networks that together perform global feature selection and compute the weights of the first hidden layer of $\mathcal{C}$ \cite{margeloiu2023weight}.
    The objectives of this addition are twofold: facilitate effective learning by $\mathcal{C}$ on HDLSS data and provide a mean to approximate the feature importance scores.
    % which are then used to inform CTGAN's training-by-sampling step such that conditions for important features are more likely to be sampled. %\hau{are you going to define CTGAN?} 
    \item Lastly, we enforce skip logic on synthetic samples during training by leveraging CTGAN's conditional vector. %\hau{of what? from where?}. %\hau{the conditional vector you mentioned earlier is for cat. data only?}
    % \item to enhance semantic integrity of the generated samples for improved utility i.e., downstream performance, we adopt the model from \cite{zhao2022ctab} (\href{https://github.com/Team-TUD/CTAB-GAN-Plus/tree/main}{code}), which also addresses the encoding of ordinal and mixed features
    % \item however, in HDLSS setting, predictions from the auxiliary classifier (AC) is likely poor and hence the downstream loss may stagnate
    % \item to tackle this, we incorporate two auxiliary networks to perform global feature selection and compute the weights of the first hidden layer of the AC \cite{margeloiu2023weight} (\href{https://github.com/andreimargeloiu/WPFS/tree/main}{code}), this would help the AC focus only on relevant features that actually play a larger role in downstream task (we don’t care if the generators produce samples whose irrelevant features are semantically incorrect!)
\end{enumerate}

% \hau{so one thing; how does each subsection ties to the above key points? also would it possible to include some sort of diagrams?}

% \hau{do you need to say what CTGAN consists of and then say something about different components? there should be some overview of the subsections and how they are put together or something}

\subsection{Conditional Generator}\label{subsec:cond-gen}

% The conditional generator in CTGAN provides an effective way to address class imbalance in categorical features.
We first revisit the two main components of CTGAN, the \emph{conditional generator} and the \emph{training-by-sampling} procedure. %, that would serve as essential building blocks for our GAN's design.
Formally, the conditional generator takes a vector $cond$ as additional input, which represents the condition ($D_{i^*}=k^*$) for some category $k^*$ of the $i^*$th categorical feature $D_{i^*}\in\{D_1,\ldots,D_{N_d}\}$ ($^*$ denotes the selected feature/category in $cond$). %($^*$ denotes the selected choice).
It is worth noting that each condition only involves \emph{one} category of \emph{one} categorical feature 
% i.e., one column in the table
and does not involve any continuous features. 
% since $cond$ was motivated in order to address class imbalance in categorical features
%, and that 
The sampling of $cond$ is twofold: sampling for the $i^*$th categorical feature and sampling for the $k^*$th category of that feature.
% \hau{you said sampled condition; what does it sample on? from the training data? according to some distributions?}
% \hau{So D1, D2, ... are the features and D1=k is a value of it? so then cond takes in a subset of D1, ..., D2 with assigned values? like D1 D2 D3 out of 10 things and then D1=2, D2=3, and D3=5?}
% \hau{you should say that that cond = (cond1, cond2, .... cond(some size)) where each condi is a binary vector of length the number of features times the total cat. of all features where condi(jk) corresponds to feature j and cat k??? }
% \hau{what about the 2 continuous features? it doens't make sense for cond?}
Let $\oplus$ denote vector concatenation e.g., given $\mathbf{x}_1=[0,0,0]$ and $\mathbf{x}_2=[1,0]$, $\mathbf{x}_1\oplus\mathbf{x}_2=[0,0,0,1,0]$. %\hau{how do you concat vectors?}.
If each categorical feature $D_i$ is encoded as a one-hot vector $\mathbf{d}_i=[\mathbf{d}_i^{(1)},\ldots,\mathbf{d}_i^{(k)},\ldots,\mathbf{d}_i^{(|D_i|)}]$,
% \bigoplus_{k=1}^{|D_i|}\mathbf{d}_i^{(k)}
where $|D_i|$ is the number of categories in $D_i$ and $\mathbf{d}_i^{(k)}\in\{0,1\}$, %\hau{what does $|Di|$ size mean here? I am not sure what dik stands for}, %[\mathbf{d}_i^{(1)},\ldots,\mathbf{d}_i^{(|D_i|)}]$.
then $cond$ is defined as $cond=\mathbf{m}_1\oplus\ldots\oplus\mathbf{m}_{N_d}$ where $\mathbf{m}_i=[\mathbf{m}_i^{(1)},\ldots,\mathbf{m}_i^{(k)},\ldots,\mathbf{m}_i^{(|D_i|)}]$ is the \emph{mask} %\hau{is mask the right word? so you sample a cond as m? what about di hat then?}
vector of zeros associated with $\mathbf{d}_i$ with $\mathbf{m}_i^{(k)}=1$ at $i=i^*$ and $k=k^*$.
% \hau{why can't you keep using di instead of mi? so this cond only has a singe 1 and everything else is 0? if that is the case, this cond is very long vector; is that an issue?}
%Intuitively, if a categorical feature $D_i$ represented by $\mathbf{d}_i$ is selected i.e., $\mathbf{d}_i=\mathbf{d}_{i^*}$ and consequently its $k$th category is chosen i.e., $k=k^{*}$ to form the condition ($D_{i^*}=k^*$), then the mask vector associated with $\mathbf{d}_i$, $\mathbf{m}_i$, will have its $k$th value set to 1.
%$cond$ is hence a vector of zeros with the entry corresponding to the sampled condition, $\mathbf{m}_{i^*}^{(k^*)}$, taking value of 1.
% \hau{what is the relationship and mi and di? i can't understand the last sentence}
To ensure the generator produces samples
% with the exact same categories as
in accordance with the given conditions, a cross-entropy term measuring the difference between $\mathbf{m}_{i^*}$ from the input $cond$ and the generated (denoted by \^{}) one-hot feature $\mathbf{\hat{d}}_{i^*}$ is added to its loss. %\hau{now there is a hat on di?} %(referred as \emph{generator loss}). 
% \hau{i think want to say cond with mi* has a 1? mi* is not an input? the input is cond with mi* component; do you need to say which part of mi* is one? i guess if you similar notation i guess you can say abuse of notations or refer to mi* instead of cond in some context?}

To help the model evenly explore all possible categories in categorical features, a procedure for sampling the $cond$ vector, termed \emph{training-by-sampling}, is employed in CTGAN as follows: randomly choose a categorical feature $D_{i^*}$ with uniform probability; % \hau{this is I at the beginning? last you update this right?}
construct the probability mass function across the categories available in $D_{i^*}$ by taking the logarithm of their frequencies in that feature (with respect to all training examples in $\mathbf{T}_{train}$); then sample a category $k^*$ accordingly and calculate the corresponding $\mathbf{m}_{i^*}$ and $cond$.
% Sampling based on the log probabilities helps increase the chance of being selected for minor categories (that exist in $\mathbf{T}_{train}$), which alleviates mode collapse in imbalanced categorical features. %\hau{if a cat doesn't show up in training, it still has some prob being selected (?)}
Afterward, the $cond$ vector is used to condition both synthetic and real training samples in order for the discriminator to properly estimate the (Wasserstein) distance between the learned and real conditional distributions $P_{\mathcal{G}}(\text{row}|cond)$ and $P(\text{row}|cond)$, respectively. %for some observation `row' from the joint distributions.
% each time we need a conditional vector during training, we first randomly choose a variable with uniform probability. Then we calculate the probability distribution of each mode (or class for categorical variables) in that variable using frequency as proxy and sample a mode based on the logarithm of its probability. Using the log probability instead of the original frequency gives minority modes/classes higher chances to appear during training. This helps to alleviate the collapse issue for rare modes/classes

% Since our goal of training generative models is to augment existing data from which classification models can improve their prediction on the target variable,
We design our GAN to place more emphasis on generating samples conditioned on $y$ by leveraging the conditional generator (discussed in the following paragraph) as well as other known techniques (discussed in subsequent subsections). %\hau{you have paragraph below? is the below new? or subsubsections?}

% \hau{the above is old stuff? the following is new stuff?}

% We discuss  shortly.
% We leverage this idea to perform two essential 

\paragraph{Proper Conditional Generation in HDLSS Setting.}
In each iteration of CTGAN, $cond$ is sampled twice during the respective training of the discriminator $\mathcal{D}$ and the generator $\mathcal{G}$, with sample size equal the specified batch size for both. %during the training of both $\mathcal{D}$ and $\mathcal{G}$..
For our HDLSS setting with $|\mathbf{T}_{train}|<200$ and over 600 columns/available conditions to consider, even when the batch size is maximally set to $|\mathbf{T}_{train}|$,
% $cond$ would not be sufficiently sampled to explore every category of all categorical features. 
% That is, 
any condition ($D_{i^*}=k^*$) would either be missing or inadequately sampled in the minibatch, and hence it would be impossible for $\mathcal{G}$ to properly learn to produce samples that preserve the input conditions. % i.e., cross-entropy loss won't converge
% One straightforward way to tackle this is to
Therefore, we increase the sample size of $cond$ and hence the number of synthetic samples to generate during the training of $\mathcal{G}$ by a factor of $q>1$ with respect to the batch size ($q=1$ in CTGAN). %\hau{did you try different q to see what is the best one?}
%\hau{i am not clear about the relationship between batch size and learning??}
Moreover, instead of randomly sampling $D_{i^*}$ while constructing $cond$, we want $\mathcal{G}$ to sample certain features more frequently than others in order to prioritize learning their conditional distributions, particularly for the target variable $y$ since our main purpose of generating synthetic samples is to help classification models improve their prediction on $y$. %\hau{why you focus y? to have similar distributions?}
Therefore, we dedicate a portion of $cond$ to conditions solely for $y$. %\hau{where did you discuss how to sample?}.
We discuss on how we sample the remaining $D_{i^*}$'s in Section \ref{subsec:wpfs}.

\subsection{Auxiliary Classifier}\label{subsec:ac}

Incorporating an auxiliary classifier $\mathcal{C}$ into GAN architecture has been shown to improve conditional generation quality for both image and tabular data \cite{odena2017conditional,park2018data}.
$\mathcal{C}$ takes either a synthetic or a real sample having its label removed as input and aims to predict that label.
Its loss, termed \emph{classification loss} \cite{park2018data}, quantifies the discrepancy between the label of a real sample and the label predicted by $\mathcal{C}$ for that same sample, which is formulated as binary and categorical cross entropy for Problems (A) and (B), respectively.
The addition of $\mathcal{C}$ also introduces an extra loss term into the loss function of $\mathcal{G}$, which has the same form as the classification loss but concerns synthetic samples instead. %\hau{so C only for the synthetic data?}
We refer to this loss as \emph{downstream loss} as in \cite{zhao2022ctab}.
Altogether, $\mathcal{C}$ is trained to learn the actual correlation between the label and the features, then teach $\mathcal{G}$ how to generate realistic samples accordingly.

% This loss quantifies the discrepancy between the label of a synthetic sample and the label predicted by $\mathcal{C}$ for that same sample and is therefore used to learn the conditional distribution $P(row|y)$.

The synthetic samples that are fed to $\mathcal{C}$ are conditioned via the $cond$ vector that is sampled earlier during the training of $\mathcal{G}$ (we iteratively train $\mathcal{D}\blacktriangleright\mathcal{G}\blacktriangleright\mathcal{C}$). %\hau{what are we sampling beforehand? the features?}.
Since we are mainly concerned with learning the conditional distribution for the target variable $P_{\mathcal{G}}(\text{row}|y)$, we only feed synthetic samples that were conditioned on $y$ to $\mathcal{C}$ when computing the downstream loss. %\hau{what does this mean?}.
The proper conditional generation step in Section \ref{subsec:cond-gen} ensures that we have sufficient amount of such samples for training $\mathcal{C}$.

\subsection{Learning Important Features}\label{subsec:wpfs}

The auxiliary classifier $\mathcal{C}$ was originally proposed to be an MLP having the same architecture as $\mathcal{D}$ \cite{park2018data}.
On HDLSS data, however, $\mathcal{C}$ is very likely to overfit, especially during the first few iterations and epochs when $\mathcal{C}$ encounters few training (either real or synthetic) examples, which in turn would negatively impact $\mathcal{G}$.
Recently, \cite{margeloiu2023weight} propose a way to overcome overfitting on HDLSS tabular classification problem by adding two auxiliary networks before the first hidden layer of some classification neural network in order to reduce the number of its learnable parameters and simultaneously perform global feature selection.
We generalize this idea by integrating these networks into $\mathcal{C}$ to prevent it from overfitting.
Furthermore, by leveraging the global importance scores $\bm{s}=[s_1,\ldots,s_N]\in(0,1)^{N}$ (higher indicates greater importance) for $N$ features, which are learned by one of the auxiliary networks, we can inform the conditional generator of important categorical features during training-by-sampling.
More specifically, while sampling the $cond$ vector, we sample features $D_i$ (excluding the target variable $y$) each with probability proportional to its important score $s_i$.
Hence, conditions for features that have significant effects on the prediction of $y$ are sampled more frequently, allowing $\mathcal{G}$ to prioritize generating synthetic samples conditioned on those features.

% \hau{ok so first you sample the features? then you construct cond from the sampled features? how many features do you usually sample?}

\subsection{Enforcing Skip Logic}\label{subsec:skl-enforce}

We first provide an intuitive explanation of how enforcing skip logic on synthetic samples can benefit the training of our GAN.
Recall that CTGAN attempts to minimize the Wasserstein distance between  $P_{\mathcal{G}}(\text{row}|cond)$ and  $P(\text{row}|cond)$, where $cond$ represents some condition ($D_{i^*}=k^*$) as defined earlier in Section \ref{subsec:cond-gen}.
% \hau{so cond = (cond1, cond2, ...,) already? or just talking about any two cond?}
Let $cond_1$ and $cond_2$ be two distinct samples of $cond$ with different sampled features e.g., $D_{1}$ ($i^*=1$) and $D_{2}$ ($i^*=2$), respectively (and hence different sampled categories).
With the presence of skip logic, if $D_{1}$ and $D_{2}$ %(representing $D_{i^*}=k^*$)
%(representing $D_{j^*}=l^*$)
both belong to the same chain imposed by another feature $D_3$ and the corresponding skip constraint $D_{3}\rightarrow\{D_{1},D_{2}\}$ is enforced such that both $D_{1},D_{2}$ are omissible, then $cond_1$ and $cond_2$ must be redefined to match the same conditional vector, $cond^*$, that satisfies such constraint i.e., ($D_1=[\text{BLANK}]$) AND ($D_2=[\text{BLANK}]$).
% $\zeta(cond_1)=\zeta(cond_2)=cond^*$ where $\zeta(\cdot)$ is a function that enforces such skip constraint on some $cond$ vector \naterevise{and $cond^*$ is the resulting conditional vector.
It follows that on real data, $P(\text{row}|cond_1)=P(\text{row}|cond_2)=P(\text{row}|cond^*)$. %\hau{dont understand this sentence}.
This implies that the corresponding synthetic samples associated with either condition should follow the same distribution $P_{\mathcal{G}}(\text{row}|cond^*)$.
Therefore, enforcing skip logic by inferring $cond^*$
% from $cond_1$ and $cond_2$
effectively reduces the search space for $P_{\mathcal{G}}$, which leads to more efficient and stable learning and hence more consistency in the quality of the generated samples. %\hau{what is the cond*?}
We empirically demonstrate this claim in Section \ref{subsec:res}.

Existing methods for enforcing column-wise constraints
% (which encapsulate more general conditions than skip constraints do)
require either creating customized transformation functions coupled with validity check or using reject sampling 
% i.e., discarding rows that violate any of the specified constraints 
\cite{patki2016synthetic}, both of which are ad hoc and  highly inefficient for large number of constrained columns.
Instead, we leverage
% take advantage of
the $cond$ vector to enforce our constraints.
% We can easily achieve this since all 26 skip constraints in our survey involve only categorical features, which are consistent with the definition of $cond$ i.e., for conditioning categorical features. %\footnote{Extension to continuous features can also be achieved by further leveraging the \emph{mode-specific normalization} method from CTGAN}.
Formally, recall from Section \ref{subsec:cond-gen} that $cond$ is constructed as $\bigoplus_{i=1}^{N_d}\mathbf{m}_i$
% is concatenated by $N_d$ zero-filled mask vectors
% $\{\mathbf{m}_1,\ldots,\mathbf{m}_{N_d}\}=\{\bigoplus_{k=1}^{|D_1|}\mathbf{m}_1^{(k)},\ldots,\bigoplus_{k=1}^{|D_{N_d}|}\mathbf{m}_{N_d}^{(k)}\}$
with $\mathbf{m}_{i^*}^{(k^*)}$ %, the $k^*$th category of the $i^*$th mask,
set to 1 and all other entries set to 0 for representing the condition ($D_{i^*}=k^*$).
Let us assume the corresponding categorical feature $D_{i^*}$ from $cond$ imposes some skip constraint $\kappa$ (e.g., $D_{i^*}=$ TB3 and $k^*=$ ``No") on a chain of features $\{D_{i'}\}_{i'\in M}$, where $M\subseteq\{1,\ldots,N_d\}$ such that $|M|$ is the size of such chain.
Let $k'_{i'}\in\{1,\ldots,|D_{i'}|\}$ be a specific category that each $D_{i'}$ takes in accordance with $\kappa$ (e.g., $D_{i'}=$ TB4 and $k'_{i'}=$ [BLANK] under TB3$=$``No").
Then, we can define $\kappa$
% (in terms of propositional logic)
as the extension of a condition as follows: 
% \hau{Are you defining M with respect to Di* = k*? what is Di* = k* is not a skip constraint or has nothing to do with it?}
\begin{align}\label{eq:sk}
\begin{split}
    \kappa &= \left[D_{i^*}\rightarrow\{D_{i'}\}_{i'\in M}\right]\\
    &= \left[ (D_{i^*}=k^*) \Rightarrow \bigwedge_{i'\in M} (D_{i'}=k'_{i'}) \right].
\end{split}
\end{align}
%\hau{are you saying if the above Di* is k* then everything in M is true or fixed to k'; but it is not clear here; because Di' must be each of the k'? the math doesn't make sense}
% where $|M|$ is the size of the chain.
Therefore, whenever applicable, we can restrict $cond\xrightarrow{\kappa}\zeta(cond)$ by reconstructing the individual mask vectors of $cond$ (with a slight abuse of notation) as
% \hau{where did you define zeta? below is zeta? connect to kappa?}
\begin{equation}\label{eq:enforce-sk}
    \mathbf{m}_{i}^{(k)}\xrightarrow{\kappa}\zeta\left(\mathbf{m}_{i}^{(k)}\right)=
    \begin{cases}
        1 & \parbox[t]{.15\textwidth}{if $i\in\{i^*,i'\}$ and $k\in\{k^*,k'_{i'}\}$,}\\
        0 & \text{otherwise}.
    \end{cases}
\end{equation}
% Let $\{\kappa_1,\ldots,\kappa_{M}\}$ be a set of $M$ skip constraints where $\kappa_l=\{\mathbf{\Tilde{m}}_1,\ldots,\mathbf{\Tilde{m}}_{N_d}\}$ with $m$ such that the  for a chain of x ($i'\subseteq\{1,\ldots,N_d\}$) features.
% If $cond$ satisfies any of the constraints, we infuse such information into $cond$ by
% \begin{equation*}
%     x.
% \end{equation*}
Note that $\kappa$ is only defined for a fixed set of $\{k^*,k'_{i'}\}$ and hence can take various forms in practice. For instance, the following is an exhaustive list of valid expressions for $\kappa=\text{TB3}\rightarrow\{\text{TB4}\}$:
{\small
\begin{itemize}
    \itemsep-0.1em
    \item[$\blacktriangleright$] TB3$=$``No" $\Rightarrow$ TB4$=$[BLANK] (omissible)
    \item[$\blacktriangleright$] TB3$=$``Yes" $\Rightarrow$ TB4$=$``Not at all"
    % \item[$\blacktriangleright$] TB3$=$``Yes" $\Rightarrow$ TB4$=$[any of the categories in TB4 (Figure \ref{fig:skl})]
    \item[$\blacktriangleright$] TB3$=$``Yes" $\Rightarrow$ TB4$=$``Less than 1 cigarettes a day''
    \item[$\blacktriangleright$] TB3$=$``Yes" $\Rightarrow$ TB4$=$``1-5 cigarettes a day''
    \item[$\blacktriangleright$] TB3$=$``Yes" $\Rightarrow$ TB4$=$``Half a pack a day''
    \item[$\blacktriangleright$] TB3$=$``Yes" $\Rightarrow$ TB4$=$``A pack or more a day''.
\end{itemize}
}

\subsection{The Complete Model}\label{subsec:complete-gan}

The complete training procedure via minibatch stochastic gradient descent (SGD) is summarized in Algorithm \ref{alg:1}. 
$q$ was previously defined in Section \ref{subsec:cond-gen}.
We denote the conditions for $y$ as $cond^{(y)}$.
$\omega\in[0,1]$ is the ratio for controlling the prevalence of such conditions in some sample of $cond$.
Let $\mathcal{I}$ be the probability mass function across the categorical features (excluding $y$) wherein the probability for selecting a feature $D_{i^*}$ is defined as its normalized feature importance score: $s_{i^*}/\sum_{i=1}^{N_d-1}s_i$ ($-1$ for excluding $y$).
% the target variable among $N_d$ categorical features).
The remaining conditions in $cond$ are sampled following the training-by-sampling procedure but with features selected according to $\mathcal{I}$ (the categories in each categorical feature are still sampled according to their log probabilities in $\mathbf{T}_{train}$ as before), for which we express as $cond_j\sim\mathcal{I}$ with a slight abuse of notation. %\hau{once you selected a feature, you select one of it cat using the training log dist?}
% The expression $cond_j\sim\mathcal{I}$ is hence used to denote sampling according to
The loss for $\mathcal{D}$, $L^{\mathcal{D}}$, is defined similarly as in CTGAN i.e., WGAN-GP loss. The loss function for $\mathcal{G}$ is    $L^{\mathcal{G}}=L^{\mathcal{G}}_{orig}+L^{\mathcal{G}}_{dstream}$
where $L^{\mathcal{G}}_{orig}$ is the original loss function for $\mathcal{G}$ in CTGAN and $L^{\mathcal{G}}_{dstream}$ is the downstream loss defined in Section \ref{subsec:ac} along with the classification loss $L^{\mathcal{C}}$.
% defined as the cross entropy between the predictions $\mathcal{C}(\mathcal{G}(\cdot))$ for (label-conditioned) synthetic samples $\mathcal{G}(\cdot)$ (with labels removed) and its corresponding labels $y$.
% The classification loss $L^{\mathcal{C}}$ for training $\mathcal{C}$ is similar to $L^{\mathcal{G}}_{dstream}$ but computed using real training samples instead.

At each iteration of an epoch, the training sequence is as follows: train the discriminator (lines 3-6), train the generator (lines 7-10), and train the auxiliary classifier while simultaneously updating $\mathcal{I}$ according to $\bm{s}$ (lines 11-13).
% Note that by following the training-by-sampling procedure, it is possible for the minibatches $B$ of real examples to not be disjoint across different iterations.

% \subsection{Evaluation Protocol}

% \hau{all of the subsections before this look a bit mathematical imprecise; i wonder if they are the norms in other literature; we probably need to find someone to check more carefully}

\begin{algorithm}[h]
    \scriptsize
    \captionsetup{font=footnotesize}
	\caption{Training Our Proposed GAN}
	\label{alg:1}
	\begin{algorithmic}[1]
            \Require Discriminator $\mathcal{D}$, generator $\mathcal{G}$, auxiliary classifier $\mathcal{C}$, constants $q$ and $\omega$
	    % \item Set $D(v)\leftarrow d^G(v,s)$ for each $v\in V$ and $i\leftarrow 0$
	    
	    % \item $i\leftarrow 0$
            \item Initialize $\mathcal{I}$ uniformly
	    \For {each minibatch $B$ of equal size from $\mathbf{T}_{train}$}
                \State \begin{varwidth}[t]{\linewidth}
                Sample noise $Z$ and $cond$ (for training $\mathcal{D}$) \par
                \quad$Z=[z_1,\ldots,z_{|B|}]$ with $z_j\sim\mathcal{N}(0,1)$ \par
                \quad$cond=[cond_1,\ldots,cond_{(1-\omega)*|B|}]$ with $cond_j\sim\mathcal{I}$
                \end{varwidth}
                \State $cond\leftarrow\zeta(cond)\oplus [cond^{(y)}]_{\times\omega|B|}$ %\hau{can you do higher than $|B|$? condy according to what? condj from I is not quite correct because you have two sample steps; first you draw the features, then from the features, you sample a cat based on the T train log dist? or based on B?}
                \State Sample $|B|$ rows from $\mathbf{T}_{train}$ such that each row $\mathbf{r}_j$ satisfies $cond_j$ 
                % \State Rearrange rows in $B$ such that each row $\mathbf{r}_j$ satisfies $cond_j$ \hau{is this always possible? what if there is only one example for cond1 and cond3?}
	        \State Perform SGD update of $\mathcal{D}$ with $L^{\mathcal{D}}(\mathcal{G}(Z,cond),B)$
                \State \begin{varwidth}[t]{\linewidth}
                Sample noise $Z$ and $cond$ (for training $\mathcal{G}$ and $\mathcal{C}$) \par
                \quad$Z=[z_1,\ldots,z_{q*|B|}]$ with $z_j\sim\mathcal{N}(0,1)$ \par
                \quad$cond=[cond_1,\ldots,cond_{q*(1-\omega)*|B|}]$ with $cond_j\sim\mathcal{I}$
                \end{varwidth}
                \State $cond\leftarrow\zeta(cond)\oplus[cond^{(y)}]_{\times q\omega|B|}$
	        \State Perform SGD update of $\mathcal{G}$ with $L^{\mathcal{G}}_{orig}(\mathcal{G}(Z,cond))$
                % \State Extract samples that are conditioned on $y$ from $\mathcal{G}(Z,cond)$
                \State \begin{varwidth}[t]{\linewidth}
                Perform SGD update of $\mathcal{G}$ with \par
                $$\quad L^{\mathcal{G}}_{dstream}\left(\mathcal{C}\left(\{\mathcal{G}(Z,cond_j)\}_{cond_j\in cond^{(y)}}\right)\right)$$
                \end{varwidth}
                \State Obtain $\bm{s}$ by computing $\mathcal{C}(B)$
                \State Update $\mathcal{I}$ according to $\bm{s}$ %returned from $\mathcal{C}$
	        \State Perform SGD update of $\mathcal{C}$ with $L^{\mathcal{C}}(\mathcal{C}(B))$
	    \EndFor
	    % \item $E' \leftarrow \{(s,v_1), (s,v_2),\ldots, (s,v_k)\}$ the set of $k$ added edges 
	    % \item Return $G'\leftarrow G(V, E\cup E')$
         \Ensure Trained $\mathcal{G}$
	\end{algorithmic} 
\end{algorithm}
% \vspace{-4em}

% \hau{in the algorithm, spell out SGD or define SGD before; also using people have a paragraph explaining the algorithm}
% \hau{maybe you want to say all these learning parameters are discussed in Section ???}

\section{Experiments}\label{sec:exp}

% We provide the methodology for evaluating our proposed GAN in Section \ref{subsec:method}. 
% In Section \ref{subsec:res}, we present the experimental results accordingly using our collected dataset.
% \hau{add an overview sentence here}

\subsection{Methodology}\label{subsec:method}

% We begin by elaborating the implementation details including the data preprocessing procedure and the considered set of hyperparameters, then proceed to describe our evaluation framework, where each evaluation metric is formally defined.
% \hau{add an overview sentence here}

% We describe the data characteristics and why it is hard to apply existing generative models on it; refer to experiment section for empirical results

% \begin{itemize}
%     \item \emph{heterogeneity}: Our dataset contains xxx numerical, xxx ordinal, and xxx categorical features (we exclude xxx text features due to ???).
%     \begin{itemize}
%         \item xxx/yyy numerical features have multimodal distributions, and xxx/yyy have non-Gaussian distributions
%         \item xxx/yyy categorical features have high cardinality ($>$zzz categories); %which leads to very sparse high-dimensional feature vectors
%         xxx/yyy of the categorical features are highly imbalanced, in which the major category appears in more than zzz\% of the rows/PWUDs
%     \end{itemize}
% \end{itemize}

\subsubsection{Implementation Details}

All experiments were conducted
% under Ubuntu 20.04 on a Linux virtual machine equipped with NVIDIA GeForce RTX 3050 Ti GPU and 12th Gen Intel(R) Core(TM) i7-12700H CPU @ 2.3GHz.
% We used
using PyTorch 1.13.1, CUDA 11.7, and scikit-learn 1.3.2.
Our implementation of the proposed GAN\footnote{\url{https://github.com/AnonyMouse3005/HDLSS-GAN}} is based on CTGAN's.
% , which is open-sourced
% \footnote{\url{https://github.com/sdv-dev/CTGAN}}.
We refer readers to our full paper for further implementation details, particularly on our data preprocessing step.

\paragraph{Hyperparameters.}

For all considered generative models, unless otherwise stated, we adopt the same specifications as in the cited original work.
We use a batch size of $|B|=30$ to train each model,
% which was tuned to be as small as possible (as using a large batch size tends to cause poorer generalization \cite{keskar2016large}) until the difference in performance is negligible.
For CTGAN, we set the pac size \cite{lin2018pacgan} to 3. 
% i.e., 3 samples in each of the $30/3=10$ pacs.
$\mathcal{C}$ is a 2-layer MLP with (256, 256) neurons and either a sigmoid (for Problem A) or a softmax (for Problem B) activation in the last layer.
Each of the two attached auxiliary networks prior to the first hidden layer of $\mathcal{C}$ is a 4-layer MLP (256, 256, 256, 256) with a tanh and a sigmoid activation in the last layer, respectively.
% To ensure sufficient learning of conditional generation from $\mathcal{G}$, 
We manually tuned $q$ and $\omega$ (via 10-fold cross validation on $\mathbf{T}_{train}$) before fixing their values to 20 and 0.5, respectively. %\hau{why? did you try other parameters?}.
Each model is trained for 100 epochs, each contains $\lfloor|\mathbf{T}_{train}|/|B|\rfloor$ iterations. 
%\nateedit{(the last iteration may work with $B$ of greater size from the remainder of $|\mathbf{T}_{train}|/|B|$)}.
The ratio for partitioning $\mathbf{T}_{real}$ is $80:20$ for $\mathbf{T}_{train}$ and $\mathbf{T}_{test}$, respectively, with a total of 100 distinct seeds.

% \hau{is your manually turned the same as cross-validation of q and omega; maybe we want a citation the last sentence about why we want to the do this such of split}

% \begin{itemize}
%     \item batch size: ideally we need larger batch size for larger sample of the conditional vectors; however, \cite{keskar2016large} stated that using a larger batch tends to cause poorer generalization; therefore, we reduce the batch size to ??? (default is 500 i.e., our entire training set) while increasing the number of conditional samples to be generated during the training of $G$
%     %increasing the number of iterations of the discriminator per generator iteration to ??? ($n_{\text{critic}}$ in original WGAN paper \cite{arjovsky2017wasserstein}, default to 1 in CTGAN)
%     \item $\lambda$ (in GP): default is 10 but maybe should be different for our HDLSS data? Should also double check implementation of the entire GP term (e.g., \texttt{pac}, \texttt{alpha})
%     \item 
% \end{itemize}

% \subsubsection{Downstream Classifiers}

% Feature selection method considered: a feature evaluator e.g., \href{https://weka.sourceforge.io/doc.dev/weka/attributeSelection/CfsSubsetEval.html}{\texttt{CfsSubsetEval}} \cite{Hall1998} followed by a search algorithm e.g., greedy, genetic algorithm, \href{https://weka.sourceforge.io/doc.dev/weka/attributeSelection/BestFirst.html}{\texttt{BestFirst}}. The dataset with reduced subset of features are then fed into one of the following classifiers:

\subsubsection{Evaluation Metrics and Framework}

\begin{figure}
%\vspace{-10mm}
    \centering
    \includegraphics[width=0.4\textwidth]{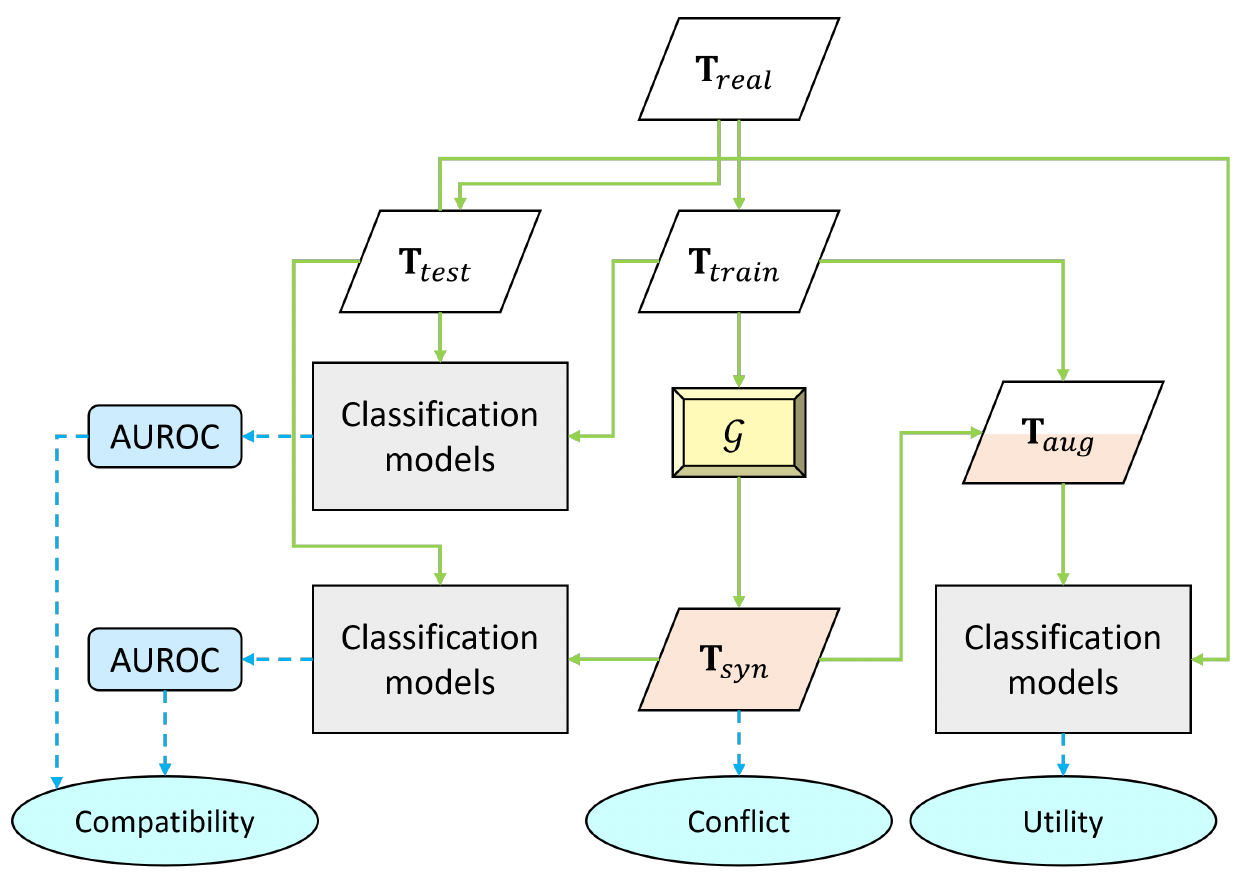}
    \caption{Workflow for evaluating the efficacy of generative models.
    The ``Classification models" blocks refer to the same set of classifiers listed in the definition of \nameref{par:compatibility}. Each of these blocks takes a table 
    % (either $\mathbf{T}_{train}$, $\mathbf{T}_{syn}$, or $\mathbf{T}_{train}+\mathbf{T}_{syn}=\mathbf{T}_{aug}$)
    as input for training the classifiers and outputs their predictions on $\mathbf{T}_{test}$. The blue dashed arrows represent the computation of the respective scores/metrics.}
    % \hau{you probably need to add captions; this picture needs a bit more explanation in text/here; there are also 3 classification models here are they all different?}
    \label{fig:eval-framework}
\end{figure}

Figure \ref{fig:eval-framework} illustrates our evaluation framework.
We evaluate the efficacy of generative models\footnote{For every measure of $\mathbf{T}_{syn}$'s quality while evaluating some trained generator $G$, we use $G$ to generate 10 samples of $\mathbf{T}_{syn}$ (each satisfies the two requirements specified in \nameref{def:tab-gen}) and evaluate their quality independently, then average the respective scores.} using three criteria: \emph{conflict},
% i.e., the degree of violation of skip logic in synthetic data, 
\emph{compatibility} \cite{park2018data},
% i.e., the discrepancy in predictive performance (on some test data) of classification models trained on synthetic data versus models trained on real data \cite{park2018data}, 
and \emph{utility}. 
% i.e., the improvement in the predictive performance of classification models when being trained on the augmented real data. % augmented by $\mathbf{T}_{syn}$.
% Although we are working with data collected from human subjects research, 
%Since all PWUDs were manually anonymized by domain experts prior to all analyses conducted in this work, we do not consider privacy preservation property. 
% such as \emph{differential privacy} as in other state-of-the-art tabular GANs \cite{jordon2018pate,park2018data}. %while working with data collected .
% We formally define these evaluation metrics as follows.
% \hau{add a sentence about we define them formally below or something}

% \hau{I think you to see the implications of each measure; what do they tell us? how do we know what is good vs bad? say ideally what do we want or something}

\paragraph{Conflict.} 
% \hau{I think you need to say where T syn is coming from; what size? what is the meaning of d head? with head?}

Every row of a synthetic table generated by $\mathcal{G}$ should not contain too many entries that violate skip logic.
Given the $j$th row of $\mathbf{T}_{syn}$ that is represented as
%\begin{equation*}
$    \mathbf{\hat{r}}_j = \mathbf{\hat{C}}_j \oplus \mathbf{\hat{d}}_{1,j}\oplus\ldots\oplus\mathbf{\hat{d}}_{N_d,j}$
%\end{equation*}
where $\mathbf{\hat{d}}_{i,j}$ is the one-hot vector of the $i$th categorical feature and $\mathbf{\hat{C}}_j$ is the representation of continuous features in that row\footnote{varied across different generative models e.g., CTGAN uses the proposed \emph{mode-specific normalization} (and so does our GAN)},
% whereas TabDDPM uses Gaussian quantile transformation \cite{scikit-learn}},
we check for each skip constraint $\kappa$ 
% \hau{did you mention how many constraints we have?}
whether the columns of $\mathbf{\hat{r}}_j$ satisfy the condition ($D_{i^*}=k^*$) i.e., match $cond$ (left-hand side of Equation \ref{eq:sk}).
If it does, the one-hot vectors of the features within the chain linked by $\kappa$ must exactly match the mask vectors associated with said features in the restricted $cond$ vector, $\zeta(\mathbf{m}_i)$ (whose construction is defined in Equation \ref{eq:enforce-sk}). Hence, we quantify the degree of $\kappa$-violation by computing the Hamming distance between the vectors $(\mathbf{\hat{r}}_{i,j})_{i\in M}=\bigoplus_{i\in M}\mathbf{\hat{d}}_{i,j}$
% $\bigoplus_{i'\in M\cup\{i^*\}}\mathbf{\hat{d}}_{i',j}$
and $\bigoplus_{i\in M}\zeta(\mathbf{m}_i)$, where $M$ contains the indices for the features within the chain linked by $\kappa$. 
% \hau{how do you set the values of mi? at the beginning it is all zero? so you don't care about cont features? also the ri are either 0 or 1 except the cont features right?}
% \hau{I am still not clear what what are M and m; why hamming distance?}
The conflict metric of a single row $\mathbf{\hat{r}}_j$ is the average Hamming distance across all applicable skip constraints, and we compute the conflict of $\mathbf{T}_{syn}$ by taking its average across all rows.
Thus, a synthetic table whose rows adequately conform skip logic yields a low conflict score in $[0,1]$.
% The conflict metric is thus defined as the average Hamming distance across all applicable skip constraints of all rows.
% \begin{equation}
    % \frac{1}{nK}\sum_{j=1}^{n}\sum_{l=1}^{K} H\left(\zeta(cond), \bigoplus_{i'\in M\cup\{i^*\}}\mathbf{\hat{d}}_{i',j} \right)
    % \sum_{l=1}^{K} \kappa_l\odot\textrm{row}
    % \frac{1}{|\mathbf{T}_{syn}|}\sum_{j=1}^{|\mathbf{T}_{syn}|} \textrm{row}\odot\zeta(cond_j)
% \end{equation}
% the (normalized) Hamming distance for each row , then 
% quantify the violation of skip constraints for a given synthetic sample, we use (normalized) Hamming distance. Normalization is necessary since the generated samples have different numbers of columns that follow skip patterns.

\paragraph{Compatibility.}\label{par:compatibility}
The classification models trained on synthetic data should output prediction for unseen examples in test data as accurately as
% achieve competitive predictive performance with respect to 
those trained on real (training) data.
We train classification models on $\mathbf{T}_{train}$ and on $\mathbf{T}_{syn}$ (both having the same size as defined in Section \ref{subsec:bg}), then test them using $\mathbf{T}_{test}$ and compare their predictive performance, which is measured using the standard Area under the Receiver Operating Characteristic curve (AUROC).
We train each of the multiclass classification models in Problem (B) using the one-vs-all strategy
% where we train a binary classifier per class (treating same-class examples as positive examples and other classes’ examples as negative examples)
and we compute the resulting AUROC also using the one-vs-all strategy
% i.e., computing the AUROC of each class against the rest 
\cite{fawcett2006introduction}.
%\hau{explain more or footnote? citations? are you classifiers also one vs all? like you have a classifier for each class? if you have many classifers how do you combine to get a single AUROC?}.

We report the average difference in AUROC (across different classification models \cite{xu2019modeling} and different partitioning of $\mathbf{T}_{train}$ and $\mathbf{T}_{test}$) of models trained on $\mathbf{T}_{syn}$ and those trained on $\mathbf{T}_{train}$. %\hau{are Train and Tyn the same size? how many time do you generate T syn since you can generate many of them right?}
% Note that we take the average AUROC across different classification models since we are not trying to determine the best one \cite{xu2019modeling}. %\hau{explain why not the best one}
It is expected for the classification models trained on $\mathbf{T}_{syn}$ to score lower AUROC than those trained on $\mathbf{T}_{train}$, ideally with a margin as small as possible.
Therefore, the compatibility score of $\mathbf{T}_{syn}$ should be negative and close to zero.
We consider the following classifiers same for both Problems (A) and (B): elastic-net logistic regression \cite{zou2005regularization},
% (with elastic-net regularization \cite{zou2005regularization}), 
decision tree i.e., CART \cite{breiman2001random}, random forest \cite{ho1995random}, XGBoost \cite{chen2016xgboost}, CatBoost \cite{prokhorenkova2018catboost}, 3-layer MLP (100, 100, 10) with sigmoid/softmax activation in the last layer, and WPFS \cite{margeloiu2023weight}. 
% \hau{which are for Problem A or Problem B?}
% \begin{itemize}
%     \item simple and interpretable
%     \begin{itemize}
%         \item regularized logistic regression (elastic-net)
%         \item decision tree i.e., CART
%     \end{itemize}
%     \item tree-ensemble e.g., XGBoost, random forest
%     \item neural networks e.g., MLP, 
% \end{itemize}

% AUROC for classification performance. We only report the highest scores returned from all considered classifiers. (Random forest and to a lesser extent XGBoost for most cases.)

\paragraph{Utility.}
% Similar to compatibility metrics, but this time
Ultimately, when the real training data is augmented by synthetic samples, the classification models trained on it should excel compared to the models trained on real data only.
We report the average AUROC (across different classification models and different partitioning of $\mathbf{T}_{train}$ and $\mathbf{T}_{test}$) of the models trained on the augmented data, $\mathbf{T}_{train}+\mathbf{T}_{syn}$, and compare it against the average AUROC of those trained on $\mathbf{T}_{train}$.
We consider the same set of classifiers listed above.
Unlike compatibility, we do not compute the difference in AUROC scores.
Hence, the utility of $\mathbf{T}_{syn}$ follows the same scale as AUROC $\in[0,1]$ and should ideally yield values as high as possible.

% \hau{so like utility is the same as compatibility; you have a model on Train for both measures but one is model is only in Ttrain and the other model is Ttrain+Tsyn}

% \hau{when you compare it; you don't show the differences right?}

% \subsubsection{Baselines}

\begin{figure}
     \centering
     % \begin{subfigure}[b]{0.5\textwidth}
     %     \centering
     %     \includegraphics[width=\textwidth,trim={2.5cm 7.7cm 0.9cm 0.2cm},clip]{images/ijcai/utility-a.pdf}
     %     % \caption{Conflict}
     %     % \label{fig:y equals x}
     % \end{subfigure}\\\medskip
     \begin{subfigure}[b]{0.48\textwidth}
         \centering
         \includegraphics[width=\textwidth]{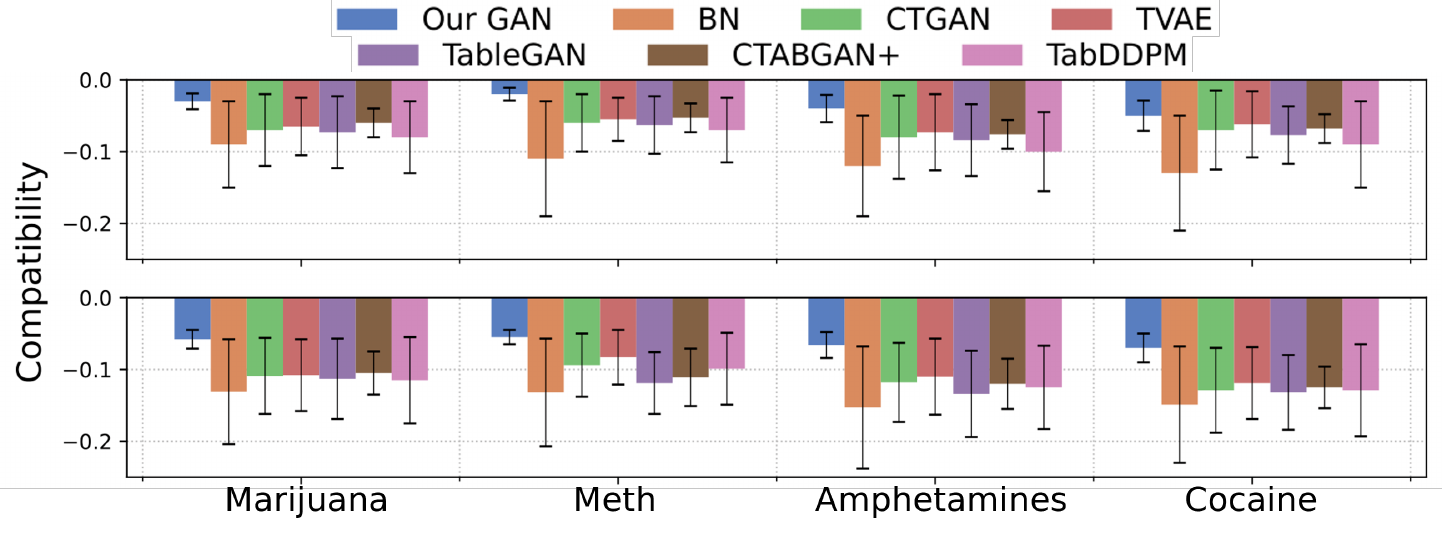}
         \caption{Compatibility (greater i.e., less negative is better)}
         \label{sfig:compatibility}
     \end{subfigure}\\
     \begin{subfigure}[b]{0.48\textwidth}
         \centering
         \includegraphics[width=\textwidth]{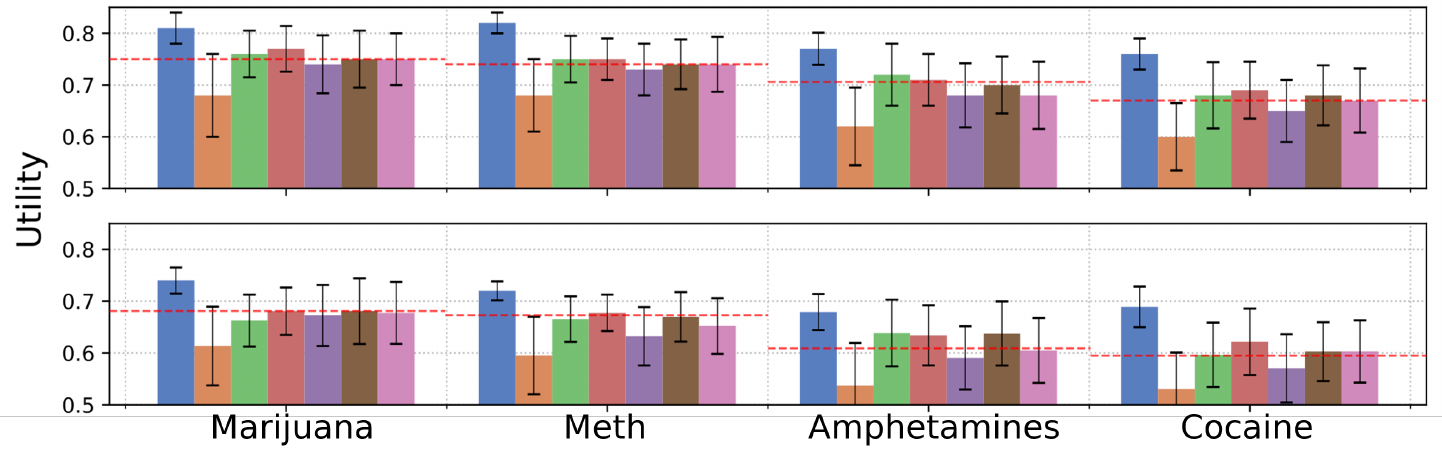}
         \caption{Utility (greater is better)}
         \label{sfig:utility}
     \end{subfigure}
        \caption{Efficacy of considered generative models in Problem (A) (top) and Problem (B) (bottom). Each column considers one drug. The red dashed lines in Figure \ref{sfig:utility} mark the average AUROC of classification models trained on $\mathbf{T}_{train}$ without data augmentation.} %\hau{top caption says higher is better? is that a typo?}}
        \label{fig:res}
\end{figure}

\subsection{Results}\label{subsec:res}

% \hau{Actually, what drugs are you showing? where are the results for other drugs? I remember you mentioned you only consider 4 but i don't see them being mentioned here in the figure or text? maybe highlight them and mention other results show similar results in the captions and and in the main texts}

% \paragraph{Tradeoff between Statistical Similarity and Preservation of Skip Logic.}
We use our evaluation framework to evaluate CTGAN, TVAE \cite{xu2019modeling}, TableGAN \cite{park2018data}, CTABGAN+ \cite{zhao2022ctab}, and TabDDPM \cite{kotelnikov2023tabddpm} in addition to the proposed GAN, which are considered state-of-the-art for tabular data generation.
For baseline, we use Bayesian networks \cite{pearl2011bayesian}.
% For baselines, we chose dummy models that can model and synthesize tabular data containing both heterogeneous types and missing values. The baselines are implemented using the open source package \texttt{synthcity} \cite{qian2023synthcity}.

% \begin{itemize}
%     \item Bayesian network
%     % \item Gaussian Copula model from \cite{patki2016synthetic}, which is based on ``copula" functions
%     \item TableGAN \cite{park2018data}
%     \item CTGAN \cite{xu2019modeling}
%     \item CTABGAN+ \cite{zhao2022ctab}
%     \item GOGGLE \cite{liu2022goggle}
%     \item TabDDPM \cite{kotelnikov2023tabddpm}
% \end{itemize}

% \paragraph{Preservation of Skip Logic.}

% \hau{maybe add paragraph below based on what results you are showing?}

\paragraph{Compatibility of Synthetic Data.}
Figure \ref{sfig:compatibility} shows the performance of all considered generative models in terms of compatibility. 
Across all considered drugs in both problems, we see that the drop in predictive performance of classifications models trained on synthetic data that are generated from our GAN ($<$0.05 and $<$0.07 lower in AUROC for Problems A and B, respectively) is considerably smaller with lower uncertainty compared to the performance drop of those trained on synthetic data from other generative models.
This demonstrates the efficacy of our GAN in generating synthetic samples that are comparable to real training data.

\paragraph{Utility of Synthetic Data.}
In terms of utility, the proposed GAN also performs well in each problem relative to the benchmarks as shown in Figure \ref{sfig:utility}. 
% \hau{you need to say a sentence about compatibility and a separate sentence about utility; and say their implications and what should we expect}
% \hau{so from the figure, you are saying that classifiers have similar performance under fake data and real data under our model? implying we are generating better examples?} 
% \hau{in the figure about utility, you mentioned that you compare to only Train; where is that being done? what it is showing? the results for both Ttrain + Tsyn?}
More specifically, the classification models trained using augmented data from our GAN gain from 8.35\% up to 13.4\% in AUROC in Problem (A) and 8.66\% up to 15.8\% in Problem (B), whereas those trained using augmented data from other models barely show any improvement.  
This implies that our GAN is capable of augmenting existing HDLSS data in order for classification models to effectively improve their predictive performance. 

% From the above observations, we can infer that there is a positive correlation between compatibility and utility scores---synthetic data with high compatibility would likely yield high utility as well and vice versa.
% Additionally, we do observe an overall performance decrease in terms of both compatibility and utility in Problem (B) with respect to Problem (A).
% One reasonable explanation could be for multiclass classification problems, the number of rows in each class are significantly less, especially for less prevalent drugs such as amphetamines and cocaine, which exacerbates the age-old problem of class imbalance.

\paragraph{Violation of Skip Logic in Synthetic Data.}
The evaluation on degree of skip logic violation is summarized in Table \ref{tab:skl}.
Since our work is the first to take this criteria into account, our proposed GAN has a clear advantage over existing models, with $\mathbf{T}_{syn}$ consistently having lower average conflict by a large margin in both considered problems.
The additional runtime for enforcing all 26 skip constraints is negligible i.e., for approximately 5\% longer training time.

\begin{table}[h]
    \centering
    \scriptsize
    \begin{tabular}{c|c|c}
        \thickhline
        Model & Problem (A) & Problem (B) \\\hline
        BN & $0.471 \pm 0.113$ & $0.490 \pm 0.154$ \\
        CTGAN & $0.339 \pm 0.074$ & $0.358 \pm 0.085$ \\
        TVAE & $0.297 \pm 0.072$ & $0.325 \pm 0.080$ \\
        TableGAN & $0.417 \pm 0.108$ & $0.452 \pm 0.094$ \\
        CTABGAN+ & $0.351 \pm 0.086$ & $0.379 \pm 0.084$ \\
        TabDDPM & $0.377 \pm 0.107$ & $0.390 \pm 0.090$ \\\hline
        \textbf{Our GAN} & $\mathbf{0.196 \pm 0.048}$ & $\mathbf{0.218 \pm 0.039}$ \\
        \thickhline
    \end{tabular}
    \caption{Conflict in $\mathbf{T}_{syn}$ (averaged over all splits of $\mathbf{T}_{real}$, lower is better). Both problems concern meth usage.}
    \label{tab:skl}
\end{table}

% \paragraph{Compatibility of Synthetic Data.}

% \paragraph{Predictive Performance.}

% add synthetic data into T train and test on T test; show tables and learning curves

% \paragraph{The Impact of Embedded Feature Selection in GAN's Classifier.}

% \subsubsection{Ablation Study}

% \hau{what is the arrow in table 3 representing? loss are lower are they better?}

\paragraph{Impact of Enforcing Skip Logic.}
We also perform an ablation study to understand the practical benefits of enforcing skip logic during our GAN training. As shown in Table \ref{tab:ablation}, when we enforce skip logic, the training phase exhibits not only higher stability but also higher efficiency, as evidenced by the loss of $\mathcal{G}$ at 50 and 100 epochs.
Note that the adoption of WGAN (via CTGAN) in our GAN allows us to interpret the loss in a meaningful way.
As a result, the trained GAN is able to consistently generate high-quality samples, which positively affects the scores for all three evaluation metrics to some extent.
Similar results for other considered problems can be found in the appendix of our full paper.
% \hau{Table 3} results don't know as good i think but at least it is better 
% \hau{say something about other drugs/problems results? in appendix?}

\begin{table}[h]
    \centering
    \scriptsize
    \begin{tabular}{c|c|c}
        \thickhline
        \multirow{2}{*}{Criteria} & \multicolumn{2}{c}{Enforcing Skip Logic?} \\\cline{2-3}
        & No & Yes \\\hline
        $L^{\mathcal{G}}_{orig}$ @50 ($\downarrow$) & $-2.276 \pm 0.849$ & $\mathbf{-3.472 \pm 0.653}$ \\
        $L^{\mathcal{G}}_{orig}$ @100 ($\downarrow$) & $-4.542 \pm 0.933$ & $\mathbf{-4.871 \pm 0.795}$ \\
        Conflict ($\downarrow$) & $0.358 \pm 0.096$ & $\mathbf{0.196 \pm 0.048}$ \\
        Compatibility ($\uparrow$) & $-0.026 \pm 0.017$ & $\mathbf{-0.023 \pm 0.009}$ \\
        Utility ($\uparrow$) & $0.818 \pm 0.038$ & $\mathbf{0.821 \pm 0.023}$ \\
        \thickhline
    \end{tabular}
    \caption{Impact of enforcing skip logic during the training of our GAN in Problem (A) for meth. The (original) loss of $\mathcal{G}$ is recorded at 50 epochs ($L^{\mathcal{G}}_{orig}$ @50) and 100 epochs ($L^{\mathcal{G}}_{orig}$ @100). ($\downarrow$) denotes less is better and ($\uparrow$) denotes greater is better.} %\hau{say others are similar? see appendix?} }
    \label{tab:ablation}
\end{table}

\section{Conclusion}\label{sec:conclu}

In this paper, using HDLSS tabular data collected by our team via a survey that employs skip logic on short-term substance use behavior, we design a novel GAN for augmenting our limited tabular data in order to help classification models accurately predict short-term substance use behaviors of PWUDs: (A) whether they would increase usage of a certain drug and (B) at which ordinal frequency they would use it within the next 12 months. %\hau{one thing in the reporting we didn't see the numerical numbers exactly; we just see the bar in Figure 2 ...}
Our evaluation results demonstrate the efficacy of the proposed GAN. %under HDLSS setting. %in all considered metrics.
% predictive performance of classification models trained using the augmented data is improved by up to.
The resulting predictions for the two defined problems can ultimately be leveraged by relevant substance use organizations as a complementary forecasting tool when determining the most appropriate resource to allocate to PWUDs that need the most help. 

\section*{Acknowledgements}
This project is supported by the National Institute of General Medical Sciences of the National Institutes of Health [P20GM130461], the Rural Drug Addiction Research Center at the University of Nebraska-Lincoln, and the National Science Foundation under grant IIS:RI \#2302999. The content is solely the responsibility of the authors and does not necessarily represent the official views of the funding agencies.

\section*{Ethics Statement}

% Although achieving great promises, we do not advise on the use of the proposed GAN for generating synthetic records on the fly, but rather for forming a decision-aid system that complements decision-making from human experts.
Due to the confidentiality agreement and IRB approval for this study, we do not have access to sensitive information such as full name and gender identity. Data will be made available to applicants (with an IRB protocol and ethical research plan) upon request.

\bibliographystyle{named}
\bibliography{main}

\begin{thebibliography}{}

\bibitem[\protect\citeauthoryear{Andersson \bgroup \em et al.\egroup }{2023}]{andersson2023inpatients}
Helle~Wessel Andersson, Mats~P Mosti, and Trond Nordfjaern.
\newblock Inpatients in substance use treatment with co-occurring psychiatric disorders: a prospective cohort study of characteristics and relapse predictors.
\newblock {\em BMC psychiatry}, 23(1):1--10, 2023.

\bibitem[\protect\citeauthoryear{Arjovsky \bgroup \em et al.\egroup }{2017}]{arjovsky2017wasserstein}
Martin Arjovsky, Soumith Chintala, and L{\'e}on Bottou.
\newblock Wasserstein generative adversarial networks.
\newblock In {\em International conference on machine learning}, pages 214--223. PMLR, 2017.

\bibitem[\protect\citeauthoryear{Borisov \bgroup \em et al.\egroup }{2022}]{borisov2022deep}
Vadim Borisov, Tobias Leemann, Kathrin Se{\ss}ler, Johannes Haug, Martin Pawelczyk, and Gjergji Kasneci.
\newblock Deep neural networks and tabular data: A survey.
\newblock {\em IEEE Transactions on Neural Networks and Learning Systems}, 2022.

\bibitem[\protect\citeauthoryear{Breiman}{2001}]{breiman2001random}
Leo Breiman.
\newblock Random forests.
\newblock {\em Machine learning}, 45:5--32, 2001.

\bibitem[\protect\citeauthoryear{Chawla \bgroup \em et al.\egroup }{2002}]{chawla2002smote}
Nitesh~V Chawla, Kevin~W Bowyer, Lawrence~O Hall, and W~Philip Kegelmeyer.
\newblock Smote: synthetic minority over-sampling technique.
\newblock {\em Journal of artificial intelligence research}, 16:321--357, 2002.

\bibitem[\protect\citeauthoryear{Chen and Guestrin}{2016}]{chen2016xgboost}
Tianqi Chen and Carlos Guestrin.
\newblock Xgboost: A scalable tree boosting system.
\newblock In {\em Proceedings of the 22nd acm sigkdd international conference on knowledge discovery and data mining}, pages 785--794, 2016.

\bibitem[\protect\citeauthoryear{Colledge-Frisby \bgroup \em et al.\egroup }{2023}]{colledge2023global}
Samantha Colledge-Frisby, Sophie Ottaviano, Paige Webb, Jason Grebely, Alice Wheeler, Evan~B Cunningham, Behzad Hajarizadeh, Janni Leung, Amy Peacock, Peter Vickerman, et~al.
\newblock Global coverage of interventions to prevent and manage drug-related harms among people who inject drugs: a systematic review.
\newblock {\em The Lancet Global Health}, 2023.

\bibitem[\protect\citeauthoryear{Couper}{2008}]{couper2008designing}
Mick~P Couper.
\newblock {\em Designing effective Web surveys.}
\newblock Cambridge University Press, 2008.

\bibitem[\protect\citeauthoryear{Dillman \bgroup \em et al.\egroup }{2014}]{dillman2014internet}
Don~A Dillman, Jolene~D Smyth, and Leah~Melani Christian.
\newblock {\em Internet, phone, mail, and mixed-mode surveys: The tailored design method}.
\newblock John Wiley \& Sons, 2014.

\bibitem[\protect\citeauthoryear{Fawcett}{2006}]{fawcett2006introduction}
Tom Fawcett.
\newblock An introduction to roc analysis.
\newblock {\em Pattern recognition letters}, 27(8):861--874, 2006.

\bibitem[\protect\citeauthoryear{Fowler}{1995}]{fowler1995improving}
Floyd~J Fowler.
\newblock {\em Improving survey questions: Design and evaluation}.
\newblock Sage, 1995.

\bibitem[\protect\citeauthoryear{Goodfellow \bgroup \em et al.\egroup }{2014}]{goodfellow2014generative}
Ian Goodfellow, Jean Pouget-Abadie, Mehdi Mirza, Bing Xu, David Warde-Farley, Sherjil Ozair, Aaron Courville, and Yoshua Bengio.
\newblock Generative adversarial nets.
\newblock {\em Advances in neural information processing systems}, 27, 2014.

\bibitem[\protect\citeauthoryear{Gulrajani \bgroup \em et al.\egroup }{2017}]{gulrajani2017improved}
Ishaan Gulrajani, Faruk Ahmed, Martin Arjovsky, Vincent Dumoulin, and Aaron~C Courville.
\newblock Improved training of wasserstein gans.
\newblock {\em Advances in neural information processing systems}, 30, 2017.

\bibitem[\protect\citeauthoryear{Heckathorn}{2014}]{heckathornRDS}
Douglas~D. Heckathorn.
\newblock {Respondent-Driven Sampling: A New Approach to the Study of Hidden Populations*}.
\newblock {\em Social Problems}, 44(2):174--199, 07 2014.

\bibitem[\protect\citeauthoryear{HHS}{2023}]{drug-report1}
HHS.
\newblock {Sidebar: The Many Consequences of Alcohol and Drug Misuse | Surgeon General’s Report on Alcohol, Drugs, and Health [Internet]}.
\newblock \url{https://addiction.surgeongeneral.gov/sidebar-many-consequences-alcohol-and-drug-misuse}, 2023.
\newblock Cited: 2023-03-18.

\bibitem[\protect\citeauthoryear{Ho}{1995}]{ho1995random}
Tin~Kam Ho.
\newblock Random decision forests.
\newblock In {\em Proceedings of 3rd international conference on document analysis and recognition}, volume~1, pages 278--282. IEEE, 1995.

\bibitem[\protect\citeauthoryear{Karamouzian \bgroup \em et al.\egroup }{2022}]{karamouzian2022latent}
Mohammad Karamouzian, Andreas Pilarinos, Kanna Hayashi, Jane~A Buxton, and Thomas Kerr.
\newblock Latent patterns of polysubstance use among people who use opioids: A systematic review.
\newblock {\em International Journal of Drug Policy}, 102:103584, 2022.

\bibitem[\protect\citeauthoryear{Kotelnikov \bgroup \em et al.\egroup }{2023}]{kotelnikov2023tabddpm}
Akim Kotelnikov, Dmitry Baranchuk, Ivan Rubachev, and Artem Babenko.
\newblock Tabddpm: Modelling tabular data with diffusion models.
\newblock In {\em International Conference on Machine Learning}, pages 17564--17579. PMLR, 2023.

\bibitem[\protect\citeauthoryear{Lin \bgroup \em et al.\egroup }{2018}]{lin2018pacgan}
Zinan Lin, Ashish Khetan, Giulia Fanti, and Sewoong Oh.
\newblock Pacgan: The power of two samples in generative adversarial networks.
\newblock {\em Advances in neural information processing systems}, 31, 2018.

\bibitem[\protect\citeauthoryear{Linden-Carmichael \bgroup \em et al.\egroup }{2022}]{linden2022stress}
Ashley~N Linden-Carmichael, Natalia Van~Doren, Bethany~C Bray, Kristina~M Jackson, and Stephanie~T Lanza.
\newblock Stress and affect as daily risk factors for substance use patterns: An application of latent class analysis for daily diary data.
\newblock {\em Prevention science}, 23(4):598--607, 2022.

\bibitem[\protect\citeauthoryear{Liu \bgroup \em et al.\egroup }{2022}]{liu2022goggle}
Tennison Liu, Zhaozhi Qian, Jeroen Berrevoets, and Mihaela van~der Schaar.
\newblock Goggle: Generative modelling for tabular data by learning relational structure.
\newblock In {\em The Eleventh International Conference on Learning Representations}, 2022.

\bibitem[\protect\citeauthoryear{Lorvick \bgroup \em et al.\egroup }{2023}]{lorvick2023protocol}
Jennifer Lorvick, Jordana Hemberg, Madeleine~J George, Joy Piontak, and Megan~L Comfort.
\newblock Protocol: Understanding polysubstance use at the daily and event levels: protocol for a mixed-methods qualitative and ecological momentary assessment study in a community-based sample of people who use illicit drugs in oakland, california, usa.
\newblock {\em BMJ Open}, 13(9), 2023.

\bibitem[\protect\citeauthoryear{Margeloiu \bgroup \em et al.\egroup }{2023}]{margeloiu2023weight}
Andrei Margeloiu, Nikola Simidjievski, Pietro Lio, and Mateja Jamnik.
\newblock Weight predictor network with feature selection for small sample tabular biomedical data.
\newblock In {\em Proceedings of the AAAI Conference on Artificial Intelligence}, volume~37, pages 9081--9089, 2023.

\bibitem[\protect\citeauthoryear{Newcomb and Bentler}{1989}]{drug-report2}
Michael~D Newcomb and Peter~M Bentler.
\newblock Substance use and abuse among children and teenagers.
\newblock {\em American psychologist}, 44(2):242, 1989.

\bibitem[\protect\citeauthoryear{{NIDA}}{2020}]{drug-report3}
{NIDA}.
\newblock {Introduction [Internet]}.
\newblock \url{https://nida.nih.gov/research-topics/commonly-used-drugs-charts}, 2020.
\newblock Cited: 2023-03-18.

\bibitem[\protect\citeauthoryear{{NIDA}}{2023}]{NIDAdeathrates}
{NIDA}.
\newblock {Drug Overdose Death Rates [Internet]}.
\newblock \url{https://nida.nih.gov/research-topics/trends-statistics/overdose-death-rates}, 7 2023.
\newblock Cited: 2023-07-11.

\bibitem[\protect\citeauthoryear{Odena \bgroup \em et al.\egroup }{2017}]{odena2017conditional}
Augustus Odena, Christopher Olah, and Jonathon Shlens.
\newblock Conditional image synthesis with auxiliary classifier gans.
\newblock In {\em International conference on machine learning}, pages 2642--2651. PMLR, 2017.

\bibitem[\protect\citeauthoryear{ODPHP}{2020}]{HHS:healthy-ppl30}
ODPHP.
\newblock Drug and alcohol use.
\newblock https://health.gov/healthypeople/objectives-and-data/browse-objectives/drug-and-alcohol-use, 2020.

\bibitem[\protect\citeauthoryear{Ouimette \bgroup \em et al.\egroup }{1997}]{ouimette1997twelve}
Paige~Crosby Ouimette, John~W Finney, and Rudolf~H Moos.
\newblock Twelve-step and cognitive-behavioral treatment for substance abuse: A comparison of treatment effectiveness.
\newblock {\em Journal of consulting and clinical psychology}, 65(2):230, 1997.

\bibitem[\protect\citeauthoryear{Park \bgroup \em et al.\egroup }{2018}]{park2018data}
Noseong Park, Mahmoud Mohammadi, Kshitij Gorde, Sushil Jajodia, Hongkyu Park, and Youngmin Kim.
\newblock Data synthesis based on generative adversarial networks.
\newblock {\em arXiv preprint arXiv:1806.03384}, 2018.

\bibitem[\protect\citeauthoryear{Patki \bgroup \em et al.\egroup }{2016}]{patki2016synthetic}
Neha Patki, Roy Wedge, and Kalyan Veeramachaneni.
\newblock The synthetic data vault.
\newblock In {\em 2016 IEEE International Conference on Data Science and Advanced Analytics (DSAA)}, pages 399--410. IEEE, 2016.

\bibitem[\protect\citeauthoryear{Pearl}{2011}]{pearl2011bayesian}
Judea Pearl.
\newblock Bayesian networks.
\newblock UCLA: Department of Statistics, 2011.

\bibitem[\protect\citeauthoryear{Prokhorenkova \bgroup \em et al.\egroup }{2018}]{prokhorenkova2018catboost}
Liudmila Prokhorenkova, Gleb Gusev, Aleksandr Vorobev, Anna~Veronika Dorogush, and Andrey Gulin.
\newblock Catboost: unbiased boosting with categorical features.
\newblock {\em Advances in neural information processing systems}, 31, 2018.

\bibitem[\protect\citeauthoryear{Ray \bgroup \em et al.\egroup }{2020}]{ray2020combined}
Lara~A Ray, Lindsay~R Meredith, Brian~D Kiluk, Justin Walthers, Kathleen~M Carroll, and Molly Magill.
\newblock Combined pharmacotherapy and cognitive behavioral therapy for adults with alcohol or substance use disorders: a systematic review and meta-analysis.
\newblock {\em JAMA network open}, 3(6):e208279--e208279, 2020.

\bibitem[\protect\citeauthoryear{Russell \bgroup \em et al.\egroup }{2021}]{russell2021qualitative}
Cayley Russell, Farihah Ali, Frishta Nafeh, Sean LeBlanc, Sameer Imtiaz, Tara Elton-Marshall, and J{\"u}rgen Rehm.
\newblock A qualitative examination of substance use service needs among people who use drugs (pwud) with treatment and service experience in ontario, canada.
\newblock {\em BMC Public Health}, 21(1):2021, 2021.

\bibitem[\protect\citeauthoryear{{SAMHSA}}{2022}]{SAMHSA2021report}
{SAMHSA}.
\newblock {Key Substance Use and Mental Health Indicators in the United States: Results from the 2021 National Survey on Drug Use and Health [Internet]}.
\newblock \url{https://www.samhsa.gov/data/sites/default/files/reports/rpt39443/2021NSDUHFFRRev010323.pdf}, 2022.
\newblock Cited: 2023-07-11.

\bibitem[\protect\citeauthoryear{Shorten and Khoshgoftaar}{2019}]{shorten2019survey}
Connor Shorten and Taghi~M Khoshgoftaar.
\newblock A survey on image data augmentation for deep learning.
\newblock {\em Journal of big data}, 6(1):1--48, 2019.

\bibitem[\protect\citeauthoryear{Street \bgroup \em et al.\egroup }{1993}]{street1993nuclear}
W~Nick Street, William~H Wolberg, and Olvi~L Mangasarian.
\newblock Nuclear feature extraction for breast tumor diagnosis.
\newblock In {\em Biomedical image processing and biomedical visualization}, volume 1905, pages 861--870. SPIE, 1993.

\bibitem[\protect\citeauthoryear{Tanner-Smith \bgroup \em et al.\egroup }{2016}]{tanner2016adolescent}
Emily~E Tanner-Smith, Katarzyna~T Steinka-Fry, Heather~H Kettrey, and Mark~W Lipsey.
\newblock {\em Adolescent substance use treatment effectiveness: A systematic review and meta-analysis}.
\newblock Peabody Research Institute, Vanderbilt University Nashville, TN, 2016.

\bibitem[\protect\citeauthoryear{\text{SAMHSA}}{2021}]{SAMHSAAdmissions}
\text{SAMHSA}.
\newblock Treatment episode data set: Admissions (teds-a).
\newblock https://www.datafiles.samhsa.gov/dataset/treatment-episode-data-set-admissions-2021-teds-2021-ds0001, 2021.

\bibitem[\protect\citeauthoryear{\text{United Nations}}{2023a}]{United-Nations:sdg3}
\text{United Nations}.
\newblock Ensure healthy lives and promote well-being for all at all ages.
\newblock https://sdgs.un.org/goals/goal3, 2023.

\bibitem[\protect\citeauthoryear{\text{United Nations}}{2023b}]{unodc2023}
\text{United Nations}.
\newblock World drug report 2023.
\newblock https://www.unodc.org/unodc/en/data-and-analysis/wdr-2023-online-segment.html, 2023.

\bibitem[\protect\citeauthoryear{Wilson and Brown}{2023}]{wilson2023barriers}
Anna~M Wilson and Aaron~R Brown.
\newblock Barriers to utilizing substance use disorder treatment and harm reduction services in appalachia.
\newblock {\em Journal of Rural Mental Health}, 2023.

\bibitem[\protect\citeauthoryear{Xu \bgroup \em et al.\egroup }{2019}]{xu2019modeling}
Lei Xu, Maria Skoularidou, Alfredo Cuesta-Infante, and Kalyan Veeramachaneni.
\newblock Modeling tabular data using conditional gan.
\newblock {\em Advances in neural information processing systems}, 32, 2019.

\bibitem[\protect\citeauthoryear{Zhao \bgroup \em et al.\egroup }{2022}]{zhao2022ctab}
Zilong Zhao, Aditya Kunar, Robert Birke, and Lydia~Y Chen.
\newblock Ctab-gan+: Enhancing tabular data synthesis.
\newblock {\em arXiv preprint arXiv:2204.00401}, 2022.

\bibitem[\protect\citeauthoryear{Zou and Hastie}{2005}]{zou2005regularization}
Hui Zou and Trevor Hastie.
\newblock Regularization and variable selection via the elastic net.
\newblock {\em Journal of the Royal Statistical Society Series B: Statistical Methodology}, 67(2):301--320, 2005.

\end{thebibliography}

\end{document}